\documentclass{article}
\usepackage{iclr2027_conference}
\iclrfinalcopy
\usepackage{newtxtext} %

\usepackage{amsmath,amsfonts,bm}

\def\eqref#1{equation~\ref{#1}}

\def\1{\bm{1}}

\DeclareMathAlphabet{\mathsfit}{\encodingdefault}{\sfdefault}{m}{sl}
\SetMathAlphabet{\mathsfit}{bold}{\encodingdefault}{\sfdefault}{bx}{n}

\newcommand{\NCells}{32}                  %
\newcommand{\NRuns}{96}                   %
\newcommand{\NSeeds}{three}
\newcommand{\NTasks}{four}
\newcommand{\NModels}{eight}

\newcommand{\EapigCellWins}{30}          %
\newcommand{\ConeVsEapig}{$8.1\times$}        %
\newcommand{\ConeVsEapigMax}{$316\times$}

\newcommand{\ConeVsEapigP}{2.5\times10^{-7}}          %
\newcommand{\ConeVsEapigCI}{$[4.9, 13.7]$}    %
\newcommand{\ConeVsEapgp}{$10.6\times$}    %
\newcommand{\EapgpCellWins}{30}            %
\newcommand{\ConeVsEapgpCI}{$[6.4, 18.1]$} %
\newcommand{\ConeVsEap}{$18.4\times$}      %
\newcommand{\ConeVsRandom}{$103\times$}    %

\newcommand{\EapCellWins}{31}

\newcommand{\ConeVsEapCI}{$[10.7, 32.9]$}
\newcommand{\RandomCellWins}{32}

\newcommand{\ConeVsRandomCI}{$[56.8, 189.7]$}
\newcommand{\AcdcCells}{8}      %
\newcommand{\AbsBarVal}{96}      %
\newcommand{\AbsBarTest}{91}     %
\newcommand{\AdaptedFullDelta}{$0.000$}  %
\newcommand{\NodesIOI}{28}\newcommand{\NodesAgr}{62}\newcommand{\NodesInd}{58}\newcommand{\NodesDoc}{150}
\newcommand{\EdgesIOI}{38}\newcommand{\EdgesAgr}{234}\newcommand{\EdgesInd}{98}\newcommand{\EdgesDoc}{497}
\newcommand{\NodesSmallest}{five}   %
\newcommand{\SftCells}{32}
\newcommand{\SftVsFrozen}{about half the size}   %
\newcommand{\SftVsCone}{$4.7\times$}             %
\newcommand{\SftVsConeWins}{28}
\newcommand{\PathTiny}{7}
\newcommand{\PathMid}{10}
\newcommand{\PathBig}{15}
\newcommand{\PathConeTiny}{$0.965$}
\newcommand{\PathFiveTiny}{$0.450$}
\newcommand{\PathConeMid}{$0.962$}
\newcommand{\PathConeBig}{$0.969$}
\newcommand{\PathFiveMid}{$0.780$}
\newcommand{\PathFiveBig}{$0.977$}
\newcommand{\PathOneMid}{$0.533$}
\newcommand{\PathOneBig}{$0.823$}
\newcommand{\PathRandMid}{$0.250$}
\newcommand{\PathRandBig}{$0.479$}
\newcommand{\PathOneTiny}{$0.026$}
\newcommand{\PathRandTiny}{$0.001$}
\newcommand{\PathFiveMatch}{21}
\newcommand{\PathOneMatch}{15}
\newcommand{\PathTieFive}{6}
\newcommand{\PathTieCone}{8}
\newcommand{\PathTieDraw}{1}
\newcommand{\PathTieMed}{$-0.001$}
\newcommand{\PathTieP}{0.79}
\newcommand{\PathKeepRatio}{$12\times$}  %
\newcommand{\ShrinkNoGemma}{$5.8\times$}       %
\newcommand{\ShrinkNoGemmaCells}{22}
\newcommand{\ShrinkAccBound}{$18.0\times$}     %
\newcommand{\ShrinkBothRestr}{$13.6\times$}    %
\newcommand{\CzeroTaskIOI}{$22.7\times$}       %
\newcommand{\CzeroTaskAgr}{$4.5\times$}
\newcommand{\CzeroTaskInd}{$12.4\times$}
\newcommand{\CzeroTaskDoc}{$2.4\times$}
\newcommand{\ShrinkIOI}{$17.7\times$}
\newcommand{\ShrinkInduction}{$18.2\times$}
\newcommand{\ShrinkAgreement}{$5.1\times$}
\newcommand{\ShrinkDoc}{$2.7\times$}
\newcommand{\ExceptionDocConeEdges}{$1{,}877$}
\newcommand{\ExceptionDocEapigEdges}{$798$}
\newcommand{\ExceptionAgreementConeEdges}{$453$}
\newcommand{\ExceptionAgreementEapigEdges}{$240$}

\newcommand{\SealedPass}{91}             %
\newcommand{\SealedShortfall}{$0.009$}
\newcommand{\SealedBoth}{71}              %
\newcommand{\SealedBothWins}{64}          %

\newcommand{\CzeroCellWins}{29}          %
\newcommand{\ConeVsCzero}{$7.4\times$}        %
\newcommand{\ConeVsCzeroCI}{$[4.3, 12.5]$}

\newcommand{\TargetShrink}{$16\times$}       %
\newcommand{\TargetShrinkMin}{$8\times$}     %
\newcommand{\TargetShrinkMax}{$128\times$}
\newcommand{\OffTargetShrink}{$1.0\times$}   %

\newcommand{\EdgeNecessityRange}{$2.5$--$15\%$}   %
\newcommand{\EdgeNecessitySmall}{$100\%$}         %
\newcommand{\EdgeNecessityFrozen}{$0.0\%$}        %
\newcommand{\MinimalCells}{19}                %
\newcommand{\MinimalMasks}{$5{,}952$}         %
\newcommand{\MinimalExact}{11}                %
\newcommand{\MinimalFrozenUntestable}{19}     %
\newcommand{\MinimalFrozenMin}{68}           %
\newcommand{\MinimalFrozenSize}{416}          %
\newcommand{\PairAblations}{$169{,}476$}  %
\newcommand{\PairCells}{48}
\newcommand{\PairPartners}{$7.4$}         %
\newcommand{\PairFrozenCostMedian}{$55\times$}  %
\newcommand{\PairFrozenCostIQR}{$7$--$759\times$}   %
\newcommand{\PairRunCells}{15}                 %
\newcommand{\AnalysisSpeedup}{$3.6\times$}

\newcommand{\RoleMatchedAcc}{$0.950$}                       %
\newcommand{\RoleCondHeads}{24}                             %
\newcommand{\RoleFrozenHeads}{61}                           %
\newcommand{\RoleCondRoles}{17}                             %
\newcommand{\RoleFrozenRoles}{25}                           %
\newcommand{\RoleFrozenUnpub}{36}                           %

\newcommand{\RolePrecGain}{$2.0\times$}                     %
\newcommand{\SwapDrop}{$0.62$}       %
\newcommand{\SwapChance}{30}         %
\newcommand{\SwapFloorCount}{27}     %
\newcommand{\SwapLlamaAcc}{$0.58$}   %
\newcommand{\SwapGemmaAcc}{$0.605$}   %
\newcommand{\RlRewardLow}{$0.008$}
\newcommand{\RlRewardHigh}{$0.95$}

\newcommand{\SftPanelPR}{$0.76$}
\newcommand{\SftPanelCells}{23}
\newcommand{\SftPanelPpl}{$1.37\times$}
\newcommand{\ContainMed}{$100\%$}
\newcommand{\ContainNull}{$8\%$}

\newcommand{\ContainTopFrac}{$13\%$}

\newcommand{\BtwoAntiCells}{32}
\newcommand{\BtwoAntiRatio}{$64\times$}   %
\newcommand{\TargetShrinkInduction}{$8\times$}            %
\newcommand{\TargetShrinkAgreement}{$4\times$}            %
\newcommand{\TargetShrinkDocstring}{$2\times$}            %
\newcommand{\RoleCondRecall}{$0.65$}                       %
\newcommand{\RoleFrozenRecall}{$0.96$}                     %
\newcommand{\RoleCondFOne}{$0.72$}                         %
\newcommand{\RoleFrozenFOne}{$0.57$}                       %
\newcommand{\RoleDropped}{9}                                %
\newcommand{\OrigAgree}{$96.5\%$}   %
\newcommand{\OrigKlGrid}{$0.215$}    %
\newcommand{\KlHeadCond}{$0.056$}                           %
\newcommand{\KlHeadFrozen}{$0.50$}                          %
\newcommand{\WangHeads}{26}
\newcommand{\HeadBudget}{20}                                %
\newcommand{\SuffCondensed}{$0.94$}                            %
\newcommand{\SuffFrozenBudget}{60}                                %
\newcommand{\SuffInductionCond}{$0.955$}
\newcommand{\SuffInductionFrozen}{$0.60$}
\newcommand{\SuffInductionBudget}{66}
\newcommand{\IIACondensed}{$0.930$}
\newcommand{\IIAFrozen}{$0.735$}
\newcommand{\IIAOriginal}{$0.645$}
\newcommand{\FailAurocCond}{$0.894$}    %
\newcommand{\FailCells}{11}            %
\newcommand{\FailWins}{10}             %
\newcommand{\FailAurocFrozen}{$0.645$}  %

\newcommand{\BoneParticipation}{$0.99$}
\newcommand{\BtwoRandom}{$50\times$}
\newcommand{\BtwoRandomMin}{$2.6\times$}

\newcommand{\SeedContainment}{$89$--$95\%$}                      %
\newcommand{\SeedNullMin}{$45\times$}                        %
\newcommand{\SeedNullMax}{$243\times$}                       %
\newcommand{\GraphMin}{$2{,}041$}
\newcommand{\GraphMax}{$43{,}993$}
\newcommand{\GraphRatio}{$22$}
\newcommand{\ScalingSlope}{$+0.10$}
\newcommand{\ScalingCI}{$[-0.21, +0.41]$}

\newcommand{\ScalingSpearman}{$+0.08$}
\newcommand{\ScalingSpearmanP}{0.50}               %

\newcommand{\WThreeCells}{32}
\newcommand{\WThreeSmaller}{5}
\newcommand{\WThreeNoConfig}{20}
\newcommand{\WThreeNoGate}{11}
\newcommand{\HolmMaxP}{2.6\times10^{-6}}   %

\usepackage{amsmath,amssymb}
\usepackage{graphicx}
\usepackage{booktabs}
\usepackage{array}
\usepackage{etoolbox}
\usepackage{wrapfig}
\AtBeginEnvironment{table}{\setlength{\belowcaptionskip}{6pt}}
\AtBeginEnvironment{table*}{\setlength{\belowcaptionskip}{6pt}}
\AtBeginEnvironment{wraptable}{\setlength{\belowcaptionskip}{6pt}}
\makeatletter
\newcommand{\captionof}[2]{\def\@captype{#1}\caption{#2}}
\makeatother
\usepackage{algorithm}
\usepackage[noend]{algpseudocode}
\usepackage{xcolor}
\usepackage{iftex}
\usepackage{hyperref}
\usepackage{url}
\ifXeTeX
  \hypersetup{pdflinkmargin=-1.5pt}
\else
  \hypersetup{pdflinkmargin=0.5pt}
\fi

\newif\ifdraftmode\draftmodefalse

\providecommand{\kl}{\mathrm{KL}}
\floatstyle{ruled}
\restylefloat{algorithm}

\title{Circuit Condensation: Post-Training that\\ Concentrates a Behavior's Causal Circuit}

\author{Sai Adith Senthil Kumar \\
George Mason University \\
\texttt{ssenthi2@gmu.edu}}

\begin{document}
\maketitle
\lhead{}

\begin{abstract}
One approach in mechanistic interpretability explains model behavior through circuits: the components and
connections that carry it. Frozen discovery often returns hundreds of edges, making them hard to
inspect, compare, or verify exhaustively. We introduce \mbox{\textbf{Circuit Condensation}}\footnote{Code: \url{https://github.com/SAI-ADITH/circuit-condensation}}, which
post-trains models to concentrate behaviors into smaller causal graphs. Each round prunes
low-attribution edges and trains a low-rank adapter to match the original through what remains. It
keeps the cut only if task accuracy and held-out perplexity remain within tolerance. Across \NTasks{}
behaviors and \NModels{} models, condensed circuits are smaller than the strongest published frozen-weight baseline
in \EapigCellWins{} of \NCells{} settings, by \ConeVsEapig{} on average and up to
\ConeVsEapigMax{}. Repeating the search without weight updates produces larger circuits in
\CzeroCellWins{} of \NCells{} settings, showing that weight updates drive the reduction rather than
search alone. For \MinimalCells{} small circuits, exhaustive subset tests show that \MinimalExact{} are exactly
minimal and identify removable edges in the rest. Pair ablations reveal dependencies that single-edge
tests miss. On indirect object identification, condensation isolates \RoleCondHeads{} heads, including
\RoleCondRoles{} with documented roles, whereas a matched frozen circuit needs \RoleFrozenHeads{}
heads, \RoleFrozenUnpub{} without documented roles. The result is a sufficient subcircuit of the
published mechanism rather than a reconstruction of it. The resulting circuit tracks the original
model's next-token distribution and, at matched edge count, predicts its errors better than a frozen
circuit in 10 of 11 evaluable settings.
\end{abstract}

\section{Introduction}\label{sec:intro}

Mechanistic interpretability seeks explanations in the internal computations that produce a language model's behavior and tests them through intervention rather than observation alone. One approach analyzes that computation as a \emph{circuit}: a subgraph of components and connections sufficient to carry a particular behavior \citep{olah2020zoom,elhage2021,olsson2022,wang2023}. The circuit view makes a precise claim that can be checked: a circuit is \emph{faithful} if the model still performs the behavior when activations outside it are replaced.

Faithfulness is only a starting point. To understand a circuit, we would also like to know whether it contains unnecessary components and how its retained edges work together. These checks quickly become combinatorial, and circuits recovered from real models are often too large to test exhaustively. Proving that no smaller circuit suffices requires testing every subset, while testing whether retained edges depend on one another requires ablating every pair. For example, \citet{wang2023} found exhaustive completeness testing infeasible for an indirect object identification (IOI) circuit containing $26$ attention heads and used three subset-sampling strategies instead. A $13$-edge circuit admits $8{,}192$ subsets and $78$ pairs; a $400$-edge circuit admits $2^{400}$ subsets. Size decides which questions can be asked at all, which is why it is a practical proxy for interpretability \mbox{\citep{gao2025}}.

Most existing methods search the fixed model more carefully. Activation and path patching exchange internal states between clean and corrupted inputs \citep{goldowskydill2023,zhangnanda2024}, Edge Attribution Patching and its integrated-gradient and saturation-corrected variants EAP-IG and EAP-GP scale that search with gradients \citep{syed2024,hanna2024,zhang2025eapgp}, and learned gates and feature-level decompositions offer further routes to sparse structure \mbox{\citep{bhaskar2024,yu2024,marks2025}}. These methods may improve the search, but they do not change how the model carries the behavior. Even within a fixed model, the recovered circuit may shift with the estimator, data, and choice among equally faithful graphs \citep{meloux2026variance,chen2026rome}.

\begin{figure}[t]
  \centering
  \includegraphics[width=\linewidth,trim=10bp 83bp 41bp 30bp,clip]{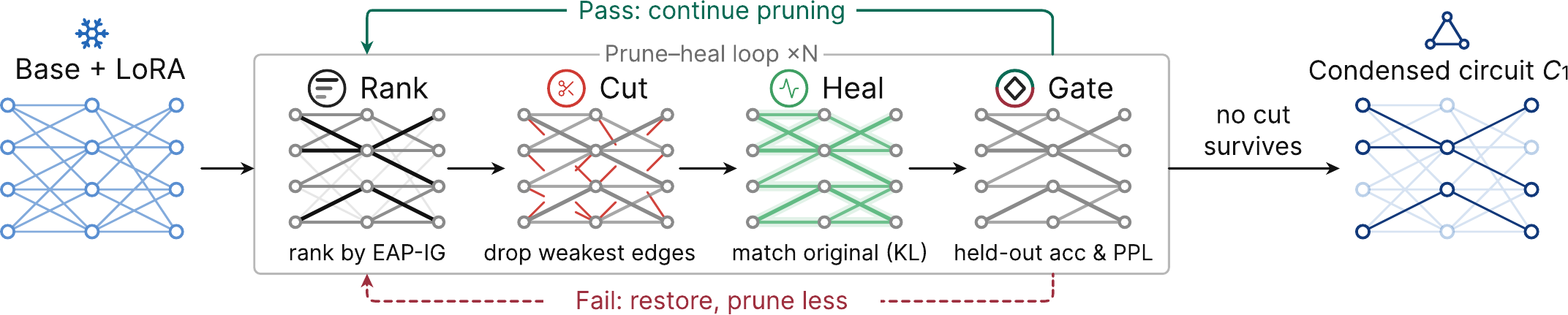}
  \caption{\textbf{One round of Circuit Condensation.} Rank surviving edges by causal
  importance, cut the weakest, then train only the adapter to reproduce the original model
  through what is left. A round is kept only if the target behavior and perplexity on held-out text remain within tolerance; otherwise the previous state is restored and a smaller cut tried. $C_1$ is the
  smallest accepted circuit.}
  \label{fig:algo-overview}
\end{figure}

We therefore ask a different question: rather than searching harder for where a behavior already lives, can training concentrate it into a smaller circuit? \textbf{Circuit Condensation} treats findability as a property a model can be trained to have. Starting from a fresh low-rank adapter \citep{hu2022lora} with every edge open, each round ranks the surviving connections by causal importance, prunes the weakest, and trains only the adapter to reproduce the original model's outputs through what remains. A controller on held-out data accepts a cut only if it preserves both the target behavior and held-out language-modeling performance, restoring the previous state otherwise (Figure~\ref{fig:algo-overview}). We call the smallest accepted circuit $C_1$.

A smaller circuit is useful only if it makes new analyses feasible and still describes the model we started from rather than only the adapted network. The reduced circuits make exhaustive subset and pairwise tests runnable (\S\ref{sec:e4}). They also track the original model's next-token distribution, at a median Kullback--Leibler (KL) divergence of $0.215$ across the grid, and predict where that model will fail better than a frozen circuit of the same size (\S\ref{sec:e5}).

\paragraph{Contributions.}
\begin{itemize}
  \item We formulate circuit findability as a trainable model property rather than a fixed target for search. Our adaptive prune, heal, and backtrack procedure concentrates a named behavior under held-out faithfulness and capability constraints and returns a validated circuit in every task--model cell without per-cell tuning.
  \item Across \NTasks{} behaviors, \NModels{} models from four families, and \NSeeds{} seeds, $C_1$ is smaller than frozen EAP-IG at comparable accuracy in \EapigCellWins{} of \NCells{} cells, by an average \ConeVsEapig{} and up to \ConeVsEapigMax{} (sign test $p=\ConeVsEapigP$, bootstrap interval \ConeVsEapigCI{}). The same search with frozen weights loses in \CzeroCellWins{} of \NCells{}: the gain is reshaping, not search.
  \item The smaller circuits enable exhaustive subset, all-pairs, and edge-level necessity checks. Matched frozen circuits would require a median \PairFrozenCostMedian{} as many pair ablations, and subset enumeration is infeasible in all \MinimalFrozenUntestable{} matched cases. We also test whether the condensed circuit still describes the unmodified model and identify where condensation does not help.
\end{itemize}

\begin{figure}[tb]
  \centering
  \includegraphics[width=\linewidth]{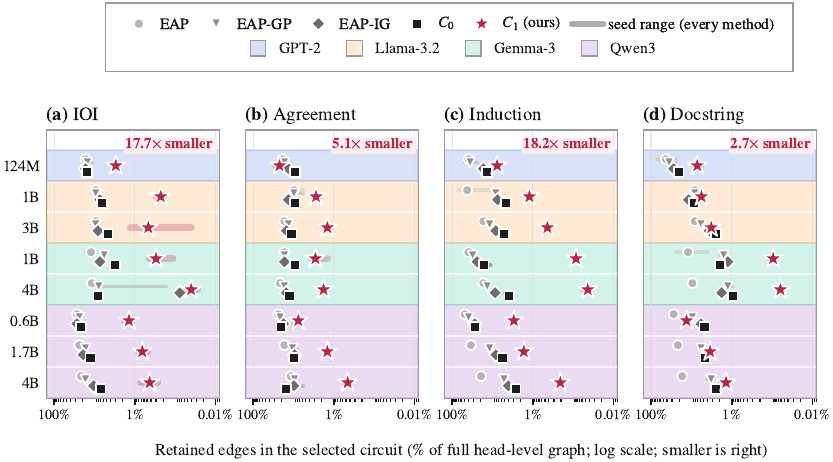}
  \caption{\textbf{Condensation usually retains fewer edges than frozen discovery at comparable
  accuracy.} Panels are behaviors and rows are models, shaded by family. The axis shows the share
  of the full head-level graph retained on a log scale, so \emph{further right means smaller}.
  Stars show $C_1$ and other markers show frozen baselines ($C_0$ runs the same search with the weights fixed); markers are means over seeds 11/22/33
  and bars span their range.}
  \label{fig:money}
\end{figure}

\section{Related Work}\label{sec:related}

\paragraph{Circuit discovery.}
Circuits explain behavior through causal subgraphs of model components \citep{olah2020zoom,elhage2021}; causal abstraction asks whether their representations respond to interventions like variables in a higher-level causal model \mbox{\citep{geiger2021,geiger2022}}. Detailed studies show what these explanations can reveal, but manual analysis does not scale readily \citep{wang2023,olsson2022,heimersheim2023}. Activation and path patching intervene on internal states \citep{goldowskydill2023,zhangnanda2024}; ACDC organizes repeated interventions into greedy pruning \citep{conmy2023}, while causal scrubbing formalizes hypothesis testing \citep{chan2022scrubbing}. EAP, EAP-IG, and EAP-GP replace repeated interventions with gradient-based edge scores \citep{syed2024,hanna2024,zhang2025eapgp}; EAP-IG generally performs best on the MIB circuit-localization track \citep{mueller2025mib}. Other methods search over information-flow routes \mbox{\citep{ferrando2024routes}}, component decompositions \mbox{\citep{hsu2024}}, position-aware edges \mbox{\citep{haklay2025}}, or learned gates \mbox{\citep{bhaskar2024,yu2024}}. All keep the model weights fixed and ask how small a faithful circuit can be recovered, making them our main comparison class.

\paragraph{What circuit discovery returns.}
Even with fixed weights, the recovered graph changes with the ablation rule, metric, data, and estimator \citep{miller2024,zhangnanda2024,meloux2026variance}, and multiple faithful graphs may describe the same behavior \citep{chen2026rome}. Circuits can also vary across datasets or model refits \citep{rai2026,bali2026headstability}, motivating work on agreement and stability \citep{li2026consistency,parekh2026circus,anani2026certified}. Feature-level methods instead build circuit nodes from learned features \citep{cunningham2023,dunefsky2024,marks2025,ameisen2025circuit,yin2026}, which other work compares with neurons and alternative representations \citep{arora2026,wu2025}. We ask whether training can make a compact circuit sufficient, changing the target rather than only the estimator.

\paragraph{Training for interpretable structure.}
Fine-tuning can preserve circuit function even as the heads involved change, but it can also make an earlier localization stale \citep{prakash2024mechanisms,tigges2024,chen2026}. It has therefore been viewed as a search over subgraphs \citep{li2025}; we optimize that movement directly. \citet{zhang2025rl} find that reinforcement-learning (RL) fine-tuning spreads a behavior's attribution mass when measured with every edge open rather than through a faithful circuit. Other methods encourage sparse weights or modular computation \citep{gao2025,liu2023}. Closest to our setting, \citet{draye2025} post-train sparse attention connectivity and report faithful task circuits up to $97\times$ smaller by edge count. Their behavior-agnostic constraint limits communication between token positions; Circuit Condensation targets the inter-component graph for one named behavior while preserving held-out task performance and general capability. Other work uses an existing circuit to restrict adaptation \mbox{\citep{prakash2025,nuraini2026}} or edits a stored association \citep{meng2022,meng2023}; we optimize the retained circuit itself. The procedure borrows iterative pruning, self-distillation, and low-rank adaptation \citep{han2015,frankle2019,hinton2015,furlanello2018,hu2022lora}, but removes causal edges rather than parameters.

\section{Method: Circuit Condensation}\label{sec:method}

\subsection{Problem Setup}

We treat a transformer as a directed graph of residual-stream writers and readers \citep{elhage2021,conmy2023,bhaskar2024}. Writers are the token embedding, each attention head's residual contribution, and each multilayer perceptron (MLP) output; readers are downstream attention and MLP inputs and the final logits. An edge $(w,r)$ exists whenever $w$ is upstream of $r$, giving $2{,}041$ candidate edges in GPT-2 up to $43{,}993$ in Qwen3-4B.

A behavior is specified by paired inputs rather than labels: a clean prompt $x$ and a matched corrupted prompt $\tilde x$, identical in form but with a different correct answer. The corrupted prompt provides a testable replacement for a removed edge: we replace its clean activation with the corresponding counterfactual activation. Each edge carries a binary gate $z_{wr}\in\{0,1\}$. Let $a_w$ and $\tilde a_w$ be writer $w$'s contributions on $x$ and $\tilde x$. The input to reader $r$ is
\[
  x_r=\sum_{w\prec r}\left[z_{wr}a_w+(1-z_{wr})\tilde a_w\right],
\]
where $w\prec r$ ranges over writers upstream of $r$. An open edge passes the clean activation. A closed edge passes the corresponding corrupted activation rather than zero, following the \emph{interchange ablation} convention. Zeroing would push activations off the data manifold, confounding ``this edge mattered'' with ``this input is unfamiliar'' \citep{miller2024}.

We want the smallest graph that preserves both the target behavior and general model capability. Let $\mathrm{acc}(z,\theta)$ denote task accuracy under edge mask $z$ and adapter parameters $\theta$, and let $\rho(\theta)$ be the adapted model's off-target perplexity divided by the base model's. Condensation then seeks
\begin{equation}\label{eq:objective}
  \min_{z,\,\theta}\ \lVert z\rVert_0
  \quad\text{s.t.}\quad
  \mathrm{acc}(z,\theta)\ \ge\ \mathrm{acc}_{\text{full}}-\varepsilon,
  \qquad \rho(\theta)\ \le\ 1+\kappa,
\end{equation}
with $\varepsilon=\kappa=0.05$ throughout. A mask satisfying the accuracy constraint is \emph{faithful}, and $\lVert z\rVert_0$ counts its retained edges. This counts graph edges rather than computation inside each retained component (\S\ref{sec:limitations}). Optimizing $\theta$ alongside the mask allows the adapter to reroute computation instead of only searching the existing graph. Each proposed cut must satisfy both constraints before it is accepted. If a behavior resists concentration, the controller therefore returns a larger valid circuit rather than a smaller circuit with damaged performance.

\subsection{The Loop}

Exact optimization is out of reach because $|E|$ binary gates define $2^{|E|}$ masks, so we approximate Equation~\ref{eq:objective} greedily (Figure~\ref{fig:algo-overview}). Beginning with every edge open and a fresh low-rank adaptation (LoRA) module \citep{hu2022lora}, each round ranks the active edges, prunes the weakest, heals the adapter against the reduced graph, then accepts or rejects the round. The adapter is not pre-trained, so its final weights come only from accepted rounds.

\paragraph{Rank.} Ranking needs a per-edge score cheap enough to recompute every round. Edge attribution patching estimates the effect of closing an edge from a gradient instead of running the ablation, and EAP-IG averages that gradient along an interpolation path \citep{syed2024,hanna2024}; we integrate in gate space between the fully corrupted and current circuits rather than between input embeddings (Appendix~\ref{app:provenance}). We use EAP-IG because the risk it addresses grows as the graph shrinks: one gradient sample can under-score an important edge in a saturated region, especially at high sparsity \mbox{\citep{zhang2025eapgp}}, who argue the interpolation path can itself pass through such regions, where the controller spends its later rounds. Scores are recomputed after each cut, reflecting what remains.

\paragraph{Prune.} The controller first proposes removing the lowest-scoring $30\%$ of active edges. If a proposal fails, it restores the last accepted state and halves the cut, down to $5\%$. Because the accuracy curve is flat and then falls sharply, one cut size would waste rounds early or overshoot late; halving on rejection shrinks the cuts as the boundary nears.

\paragraph{Heal.} Base weights stay frozen. With the mask fixed, only the adapter trains, matching the original model's output distribution rather than the task label. Writing $p_{\mathrm{orig}}$ and $p_{\mathrm{masked}}$ for the frozen model's distribution and the adapted model through the mask, healing minimizes $\kl(p_{\mathrm{orig}}(\cdot\mid x)\,\|\,p_{\mathrm{masked}}(\cdot\mid x))$. Self-distillation rather than a task loss is deliberate: a label loss would let the adapter relearn the answer through whatever path survives, giving a small circuit that no longer describes the original computation. Matching the distribution constrains outputs, not internals, so \S\ref{sec:e4} checks the mechanism for IOI, where a published circuit provides an external test.

\paragraph{Accept or restore.} Validation data are disjoint from those used for ranking and healing. A round is accepted only if both the masked circuit and the fully open adapted model meet the accuracy constraint. The fully open model must also pass the held-out off-target capability probe; otherwise both states are restored and a gentler cut is proposed. We use a hard capability gate because performance stays stable for many rounds and can collapse after a single cut. A soft penalty could trade away that collapse, whereas the gate forces the controller to undo it. The threshold is a fixed perplexity ratio, binding unevenly across families (\S\ref{sec:limitations}). Appendix~\ref{app:hyperparams} gives the schedules and Appendix~\ref{app:algorithm} the pseudocode.

\paragraph{Validation guarantee and the frozen-weight arm.} The greedy procedure does not guarantee optimality, but every returned $C_1$ satisfies both constraints on held-out validation data because a failing state is never accepted. Section~\ref{sec:e4} tests directly whether a smaller faithful subset exists for circuits that can be enumerated. Fixing $\theta$ gives $C_0$, which keeps the same controller, ranking, and stopping rule but performs no healing.

\section{Experimental Setup}\label{sec:setup}

\subsection{Tasks, Models, and Graphs}\label{sec:headlevel}

Circuit Condensation requires paired inputs that change the correct answer while preserving the prompt format. We therefore study four token-level behaviors with aligned clean and corrupted inputs: indirect object identification (IOI) \citep{wang2023}, subject-verb agreement across an attractor \citep{linzen2016}, repeated-token induction \citep{olsson2022}, and Python docstring completion \citep{heimersheim2023}. IOI also has a published head-level circuit with named functional roles, letting us test whether condensation retains a known mechanism rather than merely task accuracy.

Across these tasks, we evaluate GPT-2 small \citep{radford2019gpt2}, Llama-3.2-1B/3B-Instruct \citep{meta2024llama32}, Gemma-3-1B/4B-it \citep{gemmateam2025gemma3}, and Qwen3-0.6B/1.7B/4B \citep{yang2025qwen3}. In every model, the graph treats each attention head as a separate component and each MLP as one component. Before pruning, we verify that the fully open graph reproduces the native model across all four families. Per-layer head sums match the native block within $6\times10^{-7}$ relative error, and end-to-end logits within $6\times10^{-4}$ (Appendices~\ref{app:modelgrid} and~\ref{app:head-audit}). This check ensures that later differences arise from edge interventions rather than errors in the graph decomposition.

\subsection{Metrics and Baselines}

We measure task performance using restricted-choice accuracy at one answer position. The correct token must outrank a task-specific set of alternatives, listed in Appendix~\ref{app:answersets}, so evaluation is a deterministic comparison of logits rather than a model-graded judgment. Accuracy alone does not show whether a circuit reproduces the full model's output behavior. We therefore measure token-level Kullback--Leibler (KL) divergence, which compares their full next-token probability distributions. Across the grid, the reference is the fully open adapted model. Separate checks compare the circuit with the original model and report how often they choose the same candidate (Appendices~\ref{app:klgrid} and~\ref{app:origfaith}). We separately evaluate capability retention using perplexity on WikiText-2 \citep{merity2017wikitext} and accuracy on LAMBADA \citep{paperno2016lambada}. Neither dataset is used by the controller (Appendix~\ref{app:capability}).

The frozen comparisons are EAP, EAP-IG, EAP-GP, a random ranking, and the frozen-weight controller $C_0$ (\S\ref{sec:method}); we additionally run ACDC \citep{conmy2023} on the \AcdcCells{} cells where its cost is tractable (Appendix~\ref{app:stats}). We implement all three attribution estimators on the controller's edge gates using their published scoring equations \citep{syed2024,hanna2024,zhang2025eapgp}. Because EAP-IG and EAP-GP require adaptation to this gate space, and EAP-GP has no reference implementation, we audit the shared pipeline against exact single-edge activation patching on a tractable graph (Appendix~\ref{app:provenance}). Circuit size alone would not validate the implementation. EAP-IG serves as our primary frozen baseline.

\subsection{Data, Selection, and Reporting}

Each task--model combination uses disjoint splits of 2000 training, 2000 validation, and 1000 sealed test examples. Appendix~\ref{app:provenance} describes their construction; the capability probe is separate. Ranking and healing use only the training split. The controller evaluates each proposed circuit on the full validation split, while the sealed test split is read once, after selection.

For every arm, we select the smallest circuit whose validation accuracy is within $0.05$ of that arm's fully open graph. Each method is therefore allowed the same accuracy drop relative to its own full graph before circuit sizes are compared (Appendix~\ref{app:matchprotocol}). Because $C_1$ is measured against the adapted full graph, adaptation could in principle lower the reference accuracy and make selection easier. In practice, the adapted graph matches the original model's full-graph accuracy to a median \AdaptedFullDelta{}. Using the original model's accuracy minus the same $0.05$ still leaves all \AbsBarVal{} endpoints qualifying on validation and \AbsBarTest{} on the sealed test split. We use \emph{matched accuracy} in \S\ref{sec:e4}, where each frozen circuit is selected to match the accuracy of the corresponding $C_1$ endpoint. We report medians over seeds 11, 22, and 33. Statistical tests treat the \NCells{} task--model combinations, rather than the \NRuns{} runs, as independent because their three seeds share a model and data.

\section{Results}\label{sec:results}

We evaluate \NTasks{} behaviors on \NModels{} models with \NSeeds{} seeds, giving \NCells{} task--model combinations and \NRuns{} runs. All arms share the graph, data, metric, and selection rule from \S\ref{sec:setup}. Circuit size is the number of retained head-level edges.

\subsection{Condensation Goes Below the Frozen Discovery Floor}\label{sec:e1}

\begin{wraptable}[13]{r}{0.50\linewidth}
  \vspace{-2\baselineskip}
  \centering\small
  \setlength{\tabcolsep}{4pt}
  \caption{\textbf{Frozen arms against $C_1$.} Ratios are geometric means (frozen/$C_1$) over all \NRuns{} seeded runs. \emph{Cells} reports wins among the \NCells{} task--model combinations at the seed median. CIs resample those combinations; all five comparisons remain significant after Holm correction (Appendix~\ref{app:stats}).}
  \label{tab:arms}
  \begin{tabular}{@{}lrcl@{}}
    \toprule
    Arm & vs $C_1$ & cells & 95\% CI \\
    \midrule
    Random & \ConeVsRandom{} & \RandomCellWins{}/\NCells{} & \ConeVsRandomCI{} \\
    EAP    & \ConeVsEap{}    & \EapCellWins{}/\NCells{}    & \ConeVsEapCI{} \\
    EAP-GP & \ConeVsEapgp{}  & \EapgpCellWins{}/\NCells{}  & \ConeVsEapgpCI{} \\
    EAP-IG & \ConeVsEapig{}  & \EapigCellWins{}/\NCells{}  & \ConeVsEapigCI{} \\
    $C_0$  & \ConeVsCzero{}  & \CzeroCellWins{}/\NCells{}  & \ConeVsCzeroCI{} \\
    \bottomrule
  \end{tabular}
\end{wraptable}

\textbf{At comparable accuracy, $C_1$ retains fewer edges than frozen EAP-IG in \EapigCellWins{} of \NCells{} task--model cells at the seed median} (Figure~\ref{fig:money}). Across all \NRuns{} runs, the average gap narrows from \ConeVsRandom{} under random ordering to \ConeVsEap{} for EAP, \ConeVsEapgp{} for EAP-GP, and \ConeVsEapig{} for EAP-IG, with one EAP-IG comparison reaching \ConeVsEapigMax{}. The random arm assigns frozen edges a uniform random order and otherwise sweeps the same budgets. This progression shows that ranking improves frozen discovery, but every frozen arm remains larger on average (Table~\ref{tab:arms}).

The exceptions are docstring/Qwen3-0.6B and agreement/GPT-2. At the seed median, $C_1$ keeps \ExceptionDocConeEdges{} rather than \ExceptionDocEapigEdges{} edges in the first and \ExceptionAgreementConeEdges{} rather than \ExceptionAgreementEapigEdges{} in the second. Agreement and docstring also have the weakest average reductions, \ShrinkAgreement{} and \ShrinkDoc{}, against \ShrinkIOI{} on IOI and \ShrinkInduction{} on induction (\S\ref{sec:limitations}).

\paragraph{The endpoints hold up on unseen data.} Although validation selects among healed candidates, \SealedPass{} of \NRuns{} $C_1$ endpoints still meet the $0.05$ tolerance on the sealed test split against their own full-graph accuracy (Table~\ref{tab:klgrid}). The other five miss it by a median \SealedShortfall{}.

The reduction depends on which constraint stops the controller. On IOI and induction, where the behavior's own accuracy sets the stopping point, it averages \ShrinkAccBound{} over 16 cells; where the capability gate binds first it is smaller (\S\ref{sec:limitations}). Because that gate rarely binds on Gemma, we also report the comparison without its two models: \ShrinkNoGemma{} on average, with $C_1$ smaller in \ShrinkNoGemmaCells{} of 24 cells.

\subsection{The Gain Comes From Reshaping, Not a Better Search}\label{sec:e2}

$C_1$ and $C_0$ differ only in whether $\theta$ may move, isolating the effect of weight updates. With the same ranking, schedule, and stopping rule, $C_0$ is larger in \CzeroCellWins{} of \NCells{} cells by an average \ConeVsCzero{} (Table~\ref{tab:arms}). Moving the weights without the loop is also insufficient. DPO, GRPO, and REINFORCE \citep{rafailov2023dpo,shao2024,williams1992reinforce} are inconclusive on these short-answer tasks (Appendix~\ref{app:posttrain}), so we compare supervised fine-tuning (SFT). A standard LoRA followed by the same EAP-IG sweep yields circuits \SftVsFrozen{} of frozen circuits but still \SftVsCone{} larger than $C_1$; $C_1$ is smaller in \SftVsConeWins{} of \SftCells{} cells (Appendix~\ref{app:stats}). Fine-tuning also concentrates edge attribution under the entropy and kurtosis measures of \citet{zhang2025rl}, reducing the participation ratio---the effective number of edges carrying attribution mass---to \SftPanelPR{} of its initial value in \SftPanelCells{} of \NCells{} cells. Its probe perplexity is \SftPanelPpl{}, above the controller's $1.05$ limit. Fine-tuning thus moves circuits toward condensation but only partway, roughly halving the graph where condensation removes most of it. A stronger baseline jointly trains gates and adapter under the same objective without the iterative loop. It finds no non-empty circuit preserving both task accuracy and general capability in \WThreeNoConfig{} of \WThreeCells{} cells, including \WThreeNoGate{} outright capability-gate failures. Of the 12 valid circuits, \WThreeSmaller{} are smaller than $C_1$ (Appendix~\ref{app:joint}).

\begin{figure}[t]
  \centering
  \includegraphics[width=\linewidth]{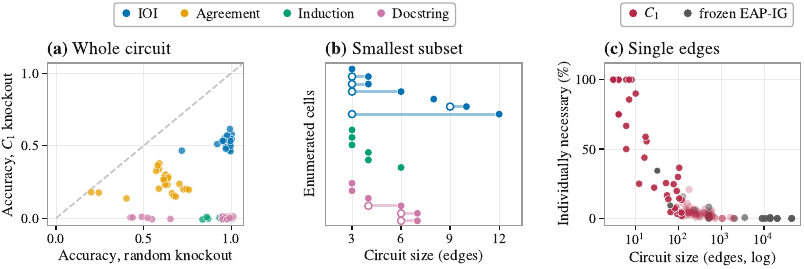}
  \caption{\textbf{Three causal checks enabled by smaller circuits.} \textbf{(a)} Accuracy after removing $C_1$ or equally many random edges. \textbf{(b)} Smallest faithful subset (open) and $C_1$ endpoint (filled), \MinimalCells{} enumerable runs. \textbf{(c)} Share of individually necessary edges; solid markers test every edge, faded ones estimate from 120 sampled edges.}
  \label{fig:verify}
\end{figure}

\subsection{The Circuit Is Small Enough to Verify}\label{sec:e4}

Smaller circuits are useful only if they still carry the behavior and make stronger causal checks tractable. We test both across the grid and compare the IOI circuit with a published mechanism.

\paragraph{The retained set carries the behavior.} Across the grid, the circuits returned by the controller range from 3 to 1{,}976 edges. For each run, we remove the entire retained set while leaving the rest of the graph open; the random control removes exactly the same number of edges. On IOI, median accuracy falls from $0.99$ to $0.52$, near the $0.50$ chance level of its two-choice metric, while random removal leaves it at $0.98$. The corresponding intact/circuit-cut/random values are $0.74/0.22/0.62$ for agreement, $1.00/0.00/0.96$ for induction, and $1.00/0.00/0.97$ for docstring (Figure~\ref{fig:verify}a; Appendix Table~\ref{tab:pertask}).

\paragraph{On IOI the circuit is a sufficient part of the published mechanism.} \citet{wang2023} assign named roles to \WangHeads{} GPT-2 heads. Frozen discovery finds \RoleFrozenRoles{} of them among \RoleFrozenHeads{} heads, so \RoleFrozenUnpub{} of its heads carry no published role. Condensation returns \RoleCondHeads{} heads, \RoleCondRoles{} named, giving a per-seed median precision gain of \RolePrecGain{} (Table~\ref{tab:roles}; Appendix~\ref{app:ioi-roles}). It keeps all three primary Name Movers and sheds three of the eight Backup Name Movers. These heads primarily take over after the primary Name Movers are ablated \citep{wang2023}, so rankings computed on the intact model may give them little weight. We then interchange the variable assigned by Wang et al.'s algorithm to the subject-inhibition heads, using the same edge indices in every arm. The intervention flips the answer as predicted in \IIACondensed{} of pairs for $C_1$, against \IIAFrozen{} for a matched frozen circuit and \IIAOriginal{} for the unmodified model; a size-matched random set scores $0.03$ and a sham hypothesis $0.00$ (Appendix~\ref{app:iia}).

\paragraph{Single-edge ablations show where slack remains.} For each circuit, we switch off every retained edge and count the fraction whose removal lowers accuracy by at least $0.05$. Across tasks, the median ranges from \EdgeNecessityRange{}. Among circuits with at most 8 edges it is \EdgeNecessitySmall{}, whereas no edge in the largest docstring circuit crosses the threshold alone. The matched frozen circuits have a median of \EdgeNecessityFrozen{} (Figure~\ref{fig:verify}c; Appendix Table~\ref{tab:pertask}). Edges below the threshold may still have smaller or joint effects.

\paragraph{Exhaustive subset tests determine exact minimality.} Single-edge tests cannot rule out a smaller faithful combination. For the \MinimalCells{} circuits small enough to enumerate, we test every subset up to $k\le13$, totaling \MinimalMasks{} masks. Eleven have no smaller faithful subset; 7 of the other 8 have only 1 to 3 removable edges. Across all 19 runs, the smallest faithful subset has 3--9 edges. Exactly minimal circuits have a median of 3 edges versus $6.5$ for those with removable edges; in the worst case, the controller keeps 12 although 3 suffice. Slack thus appears mainly in larger endpoints, consistent with stopping on a flat accuracy curve. Every matched frozen circuit exceeds the cap: the smallest has \MinimalFrozenMin{} edges and the median \MinimalFrozenSize{} (Figure~\ref{fig:verify}b; Appendix Table~\ref{tab:pertask}).

\paragraph{Small does not mean independent.} Minimality shows whether edges can be removed, not whether those remaining act independently. We ablate all $\binom{k}{2}$ pairs and compare each pair's joint effect with the sum of its individual effects. Across \PairAblations{} ablations over \PairCells{} runs, every circuit has an interacting pair. The median number of interacting partners per edge is \PairPartners{} across circuits, including in the 3-edge Gemma circuits. Condensation preserves these dependencies but makes exhaustive tests practical. Matched frozen circuits would require a median \PairFrozenCostMedian{} as many pair ablations, and only \PairRunCells{} of 32 cells have projected runtime under one day (Appendix~\ref{app:cost}). Exact single-edge patching is also \AnalysisSpeedup{} faster on the smaller graph (Appendices~\ref{app:cost} and~\ref{app:payoff-extra}).

\begin{table}[!t]
\centering\small
\setlength{\tabcolsep}{5pt}
\caption{\textbf{Shorter pruning paths fail at the smallest $C_1$ endpoints.} Every route targets $C_1$'s edge count. Each cut receives up to 2,000 healing steps, four times one controller round. \emph{Five-stage} re-ranks and heals after five cuts, \emph{single cut} ranks once and prunes directly, and \emph{random} keeps an equal-size mask. We score task accuracy only and do not apply the capability gate. Columns report seed-11 medians across \NCells{} cells by endpoint size. \emph{Matches} counts cells within $0.05$ of $C_1$ (Appendix~\ref{app:controls}).}
\label{tab:path}
\begin{tabular}{@{}lcccc@{}}
\toprule
Route & $k\!\leq\!16$ ($n{=}\PathTiny{}$) & $17$--$200$ ($n{=}\PathMid{}$) & $>\!200$ ($n{=}\PathBig{}$) & matches $C_1$ \\
\midrule
$C_1$, re-ranks every round & \PathConeTiny{} & \PathConeMid{} & \PathConeBig{} & --- \\
Five-stage cut & \PathFiveTiny{} & \PathFiveMid{} & \PathFiveBig{} & \PathFiveMatch{} \\
Single cut & \PathOneTiny{} & \PathOneMid{} & \PathOneBig{} & \PathOneMatch{} \\
Random mask & \PathRandTiny{} & \PathRandMid{} & \PathRandBig{} & $0$ \\
\bottomrule
\end{tabular}
\end{table}

\subsection{The Circuit Describes the Model We Started From}\label{sec:e5}

By construction, $C_1$ preserves the target behavior after condensation. We next ask whether its edges were already important in the original model. We rank the original edges with EAP-IG and select a frozen circuit by the same rule. It contains a median \ContainMed{} of $C_1$'s edges versus \ContainNull{} for an equal-size random set, with above-baseline overlap in all \NCells{} cells. Including every $C_1$ edge requires a median \ContainTopFrac{} of the ranked list (Appendix~\ref{app:containment}). Condensation therefore concentrates behavior into edges that frozen discovery already ranks highly. Because their weights moved, overlap alone cannot show that the computation is unchanged. Across \NRuns{} runs, $C_1$ selects the same answer as the original model on a median \OrigAgree{} of sealed-test examples, with median token-level KL \OrigKlGrid{}. On IOI/GPT-2, its KL is \KlHeadCond{} versus \KlHeadFrozen{} for the matched frozen circuit (Appendix~\ref{app:origfaith}). At equal edge count, $C_1$ predicts original-model errors with median AUROC \FailAurocCond{} versus \FailAurocFrozen{}, leading in \FailWins{} of \FailCells{} evaluable cells, all on agreement or induction (Appendix~\ref{app:failpred}).

\subsection{Controls}\label{sec:controls}

\begin{table}[t]
\centering\small
\setlength{\tabcolsep}{1.5pt}
\caption{\textbf{Controls}, in discussion order. Each row changes one ingredient; \emph{Reference} gives its comparison. Outcomes are medians over runs. The target-test row reports IOI (other targets shrink $2$--$8\times$). PR is the EAP-IG participation ratio after versus before healing (Appendix~\ref{app:controls}).}
\label{tab:controls-summary}
\begin{tabular*}{\linewidth}{@{\extracolsep{\fill}}llccll@{}}
\toprule
Control & What changes & Cells & Seeds & Outcome & Reference \\
\midrule
Shorter path (Tab.~\ref{tab:path}) & route to the endpoint & \NCells{} & 1 & 0/\PathTiny{} match at $k\le16$ & $C_1$ at equal size \\
Random order & edge ranking & \NCells{} & 3 & \BtwoRandom{} larger & $C_1$ endpoint \\
Reverse order & edge ranking & \BtwoAntiCells{} & 1 & 0 edges removed & full graph \\
No-prune heal & pruning removed & \NCells{} & 3 & PR ratio \BoneParticipation{} & before healing \\
Target test & behavior condensed & \NCells{} & 3 & target \TargetShrink{}, rest \OffTargetShrink{} & untouched model \\
Sibling swap & head identity & \NCells{} & 3 & \SwapChance{}/\NCells{} to chance & intact circuit \\
\bottomrule
\end{tabular*}
\end{table}

\paragraph{The pruning path matters.} We test three shorter routes at $C_1$'s final edge count, allowing up to 2,000 healing steps per cut. None matches $C_1$ at endpoints of 16 edges or fewer: $C_1$ retains \PathConeTiny{} accuracy versus \PathFiveTiny{} for the five-stage cut, \PathOneTiny{} for the single cut, and \PathRandTiny{} for the random mask (Table~\ref{tab:path}). The random mask matches none of the \NCells{} cells. The single cut matches it in \PathOneMatch{}, but those cells retain \PathKeepRatio{} more of the original graph than cells where it fails. Above 200 edges, the five-stage cut ties $C_1$: it wins \PathTieFive{} of \PathBig{} cells, $C_1$ wins \PathTieCone{}, and \PathTieDraw{} ties (median difference \PathTieMed{}, sign test $p=\PathTieP{}$). All three know $C_1$'s edge count in advance; the controller must discover where to stop.

\paragraph{Attribution ranking is necessary for deep pruning.} With the controller fixed, we use random or reverse edge rankings. Random ordering returns circuits a median \BtwoRandom{} larger than $C_1$ and no closer than \BtwoRandomMin{}. Reverse ordering accepts no cuts, returning the starting graph at a median \BtwoAntiRatio{} the size of $C_1$ across \BtwoAntiCells{} cells. Both preserve accuracy only by keeping most of the graph (Appendix~\ref{app:controls}).

\paragraph{Healing alone does not concentrate attribution.} With every edge open, we heal the model and measure the EAP-IG participation ratio; higher values mean attribution is spread across more edges. Its median end-to-start ratio is 0.99, and 92 of 96 runs remain within $15\%$ of their starting value. The reduction therefore requires pruning and healing together (Appendix~\ref{app:b1}).

\paragraph{Concentration remains tied to the target behavior.} A generic sparsity effect would also simplify unrelated behaviors. We condense each behavior, then apply the same frozen search to all four at matched accuracy in the adapted and original models. Only the target moves: across eight models and three seeds, IOI becomes a median \TargetShrink{} easier to localize (\TargetShrinkMin{}--\TargetShrinkMax{}), induction \TargetShrinkInduction{}, agreement \TargetShrinkAgreement{}, and docstring \TargetShrinkDocstring{}. All twelve off-target directions remain at \OffTargetShrink{} (Appendix~\ref{app:controls}).

\paragraph{The circuit depends on which heads feed the output.} A small circuit could still be arbitrary if another head in the same layer worked as well. We replace each selected head feeding the final output with another from its layer. Median accuracy falls by \SwapDrop{}, leaving \SwapChance{} of \NCells{} circuits no better than chance, including all seven endpoints with at most 16 edges. Adding a random edge leaves accuracy unchanged, so replacing the selected heads causes the collapse (Appendix~\ref{app:b4}).

\section{Limitations}\label{sec:limitations}

We test four token-level behaviors on fixed checkpoints up to 4B parameters and three seeds, so results may not transfer to other checkpoints, distributions, or graph granularities. $C_1$ describes the adapted model and closely, but not necessarily exactly, tracks the original. Edge counts depend on interchange ablation and exclude computation within retained components and the LoRA adapter. Exact minimality is feasible for only 19 endpoints. Three seeds test stability, not uniqueness, while flat accuracy curves can make the fixed schedule stop above the minimum. The IOI circuit is sufficient without reconstructing the full published mechanism. Condensation requires training and raises unseen WikiText-2 perplexity by a median $9\%$ (Appendix~\ref{app:limitations}).

\section{Conclusion}\label{sec:conclusion}

Circuit Condensation turns circuit findability from a fixed-model property into a training target. Across \NCells{} settings, it returns circuits \ConeVsEapig{} smaller than the strongest frozen baseline in \EapigCellWins{}. The resulting circuits make exhaustive subset and pair tests feasible while remaining concentrated in pathways that frozen discovery ranks highly in the original model. This makes causal analysis cheaper to repeat and expands what can be verified directly.

\subsubsection*{Acknowledgments}
This work used GPU resources on the Hopper cluster, provided by the Office of Research Computing at George Mason University (URL: \url{https://orc.gmu.edu}). The cluster is funded in part by grants from the National Science Foundation (Award Number 2018631).

\subsubsection*{Reproducibility Statement}
Appendix~\ref{app:modelgrid} documents the model checkpoints and graph construction. Appendices~\ref{app:hyperparams} and~\ref{app:provenance} give the controller settings, per-run configurations, prompt generation, data splits, scoring, and selection rule. Each run uses one 40GB or 80GB NVIDIA A100 GPU and records its seed, launch settings, pruning history, final mask, LoRA adapter, and validation and test outputs. We manually verified the reported results against the saved JSON outputs and individual runs. Code and data-generation scripts are available at \url{https://github.com/SAI-ADITH/circuit-condensation}.

\subsubsection*{AI Use Statement}
We used generative AI tools to assist with code implementation and debugging, experiment orchestration, consistency checks, figure and table preparation, literature searches, suggestions on experimental controls and interpretation, and suggestions for drafting and revising prose to improve clarity, flow, and professional presentation. The authors designed the experiments, generated and evaluated all outputs, interpreted the results, verified the experiments and reported findings, checked citations against original sources, and wrote and approved the final manuscript. Experimental examples were produced by deterministic scripts rather than directly by an LLM, and model outputs were scored without LLM grading. AI-assisted code was manually reviewed and tested, and reported results were manually verified against saved JSON outputs and individual runs. We take responsibility for all text, code, analyses, and conclusions.

\bibliography{references}
\bibliographystyle{iclr2027_conference}

\appendix
\raggedbottom
\section{Models, Graphs, and Head Decomposition}\label{app:modelgrid}

{\raggedright\noindent
We test \NTasks{} behaviors on \NModels{} models: GPT-2 small, Llama-3.2-1B/3B-Instruct, \mbox{Gemma-3-1B/4B-it}, and Qwen3-0.6B/1.7B/4B. The checkpoints are openai-community/\allowbreak{}gpt2, meta-llama/\allowbreak{}Llama-3.2-1B-Instruct, meta-llama/\allowbreak{}Llama-3.2-3B-Instruct, google/\allowbreak{}gemma-3-1b-it, google/\allowbreak{}gemma-3-4b-it, Qwen/\allowbreak{}Qwen3-0.6B, Qwen/\allowbreak{}Qwen3-1.7B, and Qwen/\allowbreak{}Qwen3-4B. Each attention head is a separate graph writer; each MLP is one writer. The candidate graph grows from \GraphMin{} edges on GPT-2 to \GraphMax{} on Qwen3-4B (\GraphRatio{} times larger). Table~\ref{tab:endpoints} reports every endpoint. We check the graph decomposition against each model's unmodified forward pass before using it for circuit experiments.\par}

\subsection{Head Decomposition Audit}\label{app:head-audit}

For each architecture, we split the native attention output before the residual addition and apply the model's own output projection to each head slice. This preserves grouped-query attention and rotary position embeddings \citep{ainslie2023,su2024rope}, architecture-specific head dimensions, and any LoRA module attached to the projection. Gemma-3 applies root mean square layer normalization (RMSNorm) after summing the attention output \citep{zhang2019rmsnorm}. We therefore compute its shared position-wise scale from the complete sum and apply it to each head contribution, so the scaled contributions add to the native normalized output. We test every architecture with both zero and nonzero adapters. Per-layer head sums match the native block within $6\times10^{-7}$ relative error, fully open end-to-end logits match within $6\times10^{-4}$, and random masks and EAP-IG passes produce no non-finite values.

\section{Controller Configuration and Pseudocode}\label{app:hyperparams}

Algorithm~\ref{alg:condensation} gives the controller used in every adaptive run. It ranks active edges, cuts the lowest-ranked ones, trains the adapter, and restores the previous state if accuracy or capability fails. We report the smallest accepted circuit; later rounds only show where further cuts fail (Table~\ref{tab:controller-hparams}).

\subsection{Run Configuration}\label{app:runconfig}

Table~\ref{tab:controller-hparams} lists the settings shared across the grid. Controls change them only where stated.

\begin{table}[H]
\centering\small
\caption{\textbf{Controller and run configuration.} Values are shared across all \NRuns{} runs and
\NCells{} cells unless a control states otherwise.}
\label{tab:controller-hparams}
\begin{tabular}{@{}l p{0.52\linewidth}@{}}
\toprule
Setting & Value \\
\midrule
\multicolumn{2}{@{}l}{\emph{Data}} \\
Data per task and model & 2000 train / 2000 validation (\texttt{eval\_big.jsonl}) / 1000 sealed test \\
Validation read per controller decision & full 2000-example split \\
Paper seeds & 11, 22, 33 \\
\midrule
\multicolumn{2}{@{}l}{\emph{Ranking and pruning}} \\
Edge-ranking method & EAP-IG, re-ranked after every accepted cut \\
Integrated-gradient steps & 5 interpolation points \\
Ranking / validation batches per round & 8 / 250 \\
Initial proposed cut & $30\%$ of active edges \\
\midrule
\multicolumn{2}{@{}l}{\emph{Healing}} \\
Healing objective & $\kl(p_{\mathrm{orig}}\|p_{\mathrm{masked}})$ on clean inputs \\
Healing steps per proposed cut & up to 500; checked every 40 steps and stopped once circuit accuracy clears the floor by a buffer (EMA of recent per-cut drops, seeded at 0.5$\times$ the tolerance, floored at 0.005) \\
Optimizer / learning rate & AdamW \citep{loshchilov2019adamw} / $10^{-4}$ \\
Trainable parameters & LoRA adapter only; base model frozen \\
LoRA configuration & rank 16, $\alpha=32$, dropout 0, no bias terms \\
\midrule
\multicolumn{2}{@{}l}{\emph{Acceptance}} \\
Cut on rejection & halved, floored at $0.05$ of active edges \\
Re-heal attempts once the cut reaches its floor & 3 \\
Overshoot rounds past the endpoint & 4 \\
Mid-heal abort check & every 40 steps \\
Adapter weight decay & $0.01$ \\
Healing / evaluation batch size & 16 / 8 \\
Edge floor (smallest returnable circuit) & 1 \\
Gate on the frozen arm ($C_0$) & inert (calibration is degenerate when no weight moves) \\
Capability-probe batches & 50 \\
Numerical precision & \texttt{bfloat16} \\
Accept accuracy tolerance & within $0.05$ of the full-circuit validation baseline, tested on a
one-sided $95\%$ lower bound ($z=1.645$) \\
Capability probe & 320 prompts across five domains \\
Capability gate / mid-heal abort & perplexity ratio $\leq 1.05$ / $>2.0$ \\
\midrule
\multicolumn{2}{@{}l}{\emph{Reporting}} \\
Arm selection rule & smallest circuit within $0.05$ of \emph{that arm's own} full-graph validation accuracy \\
Reported endpoint & smallest accepted faithful circuit, scored once on sealed test \\
Hardware & one NVIDIA A100 GPU per run (40GB or 80GB) \\
\bottomrule
\end{tabular}
\end{table}

\newpage
\subsection{Controller Pseudocode}\label{app:algorithm}

Algorithm~\ref{alg:condensation} expands the loop shown in Figure~\ref{fig:algo-overview}.

\begin{algorithm}[H]
\caption{Circuit Condensation controller}
\label{alg:condensation}
\footnotesize
\begin{algorithmic}[1]
\Require Frozen model $M$; train pairs $\mathcal{D}_{\mathrm{tr}}$; validation data $\mathcal{D}_{\mathrm{val}}$; capability probe $\mathcal{D}_{\mathrm{cap}}$
\Require Open mask $z$; initial cut fraction $q$; tolerances $\epsilon=\kappa=0.05$
\State Initialize LoRA parameters $\theta$; save $(z,\theta)$ as the accepted state
\Repeat
 \State Score active edges with EAP-IG on $\mathcal{D}_{\mathrm{tr}}$
 \State Propose $z'$ by removing the lowest-ranked fraction $q$ of active edges
 \State Optimize $\theta'$ for up to 500 steps with $z'$ fixed to minimize $\kl(p_M\|p_{M,z',\theta'})$
 \State Evaluate the masked circuit and full adapted model on $\mathcal{D}_{\mathrm{val}}$
 \State Measure the adapter-on/off perplexity ratio on $\mathcal{D}_{\mathrm{cap}}$
 \If{both target accuracies are within $\epsilon$ of baseline and the ratio is $\leq1+\kappa$}
 \State Accept and save $(z',\theta')$; continue from the smaller graph
 \Else
 \State Restore the accepted $(z,\theta)$; reduce $q$
 \EndIf
\Until{no smaller proposal can be accepted}
\State \Return the smallest accepted mask and adapter, $C_1$
\end{algorithmic}
\end{algorithm}

\section{Selection Protocol and Data Provenance}\label{app:provenance}

\paragraph{Clean and corrupted pairs.} When an edge is closed, it reads an activation from a corrupted prompt. Each clean prompt therefore has a token-aligned counterpart that changes the answer while preserving the task's structure.

\emph{IOI.} Clean prompts place the indirect object and subject in either order (ABBA and BABA templates). The corruption keeps the first clause but replaces the subject's second occurrence, the one performing the action, with a third name. The resulting prompt names three people and no longer implies the clean answer.

\emph{Agreement.} We change the head noun's number and every slot that agrees with it. We also change the attractor so it stays opposite in number to the head; the correct verb changes, but the attractor structure does not. Prompts use three frames from the agreement literature.

\emph{Induction.} The bigram $A\,B$ appears four times before a final cue $A$ (\texttt{--repeats 4}). A 6-token prefix and 6-token fillers separate repetitions, drawn from 50 single-token words. The corruption replaces every $B$ with a third word $C$, keeping the repeated pattern while changing its target.

\emph{Docstring.} We rename parameters to change the in-order copy target without changing the function shape. All substituted words are single tokens under the model's tokenizer; the agreement and docstring generators discard pairs whose tokenized lengths differ. The generators are \texttt{scripts/prepare\_\{ioi,agreement,induction,docstring\}.py}, and \texttt{scripts/prepare\_bigval.py} makes the 2000-example validation split. We pass split sizes explicitly because the scripts default to smaller samples.

\paragraph{Splits and capability probe.} We generate prompts separately for each tokenizer because substituted names and answer choices must remain single tokens. Within a task--model combination, the splits draw from the same template and lexical pools but share no clean prompts, and every arm uses the same data. Generation uses seed 0 for training, 2 for the sealed test split, and 777 for the 2000-example validation split, which excludes prompts from the existing splits. The fixed capability probe contains 320 prompts: 80 each from docstring code, IOI-style text, and GSM8K math \citep{cobbe2021gsm8k}, plus 40 each from general prose and a history task. Task-shaped prompts come from training splits. The probe is sampled once with seed 0 and reused across runs, and every run checks that it overlaps with neither validation nor test data. WikiText-2 and LAMBADA are separate evaluations that never enter the controller. The probe builder is \texttt{scripts/build\_probe\_corpus.py}.

\paragraph{Frozen-estimator implementation audit.} EAP, EAP-IG, and EAP-GP score the edge gates in our graph. EAP uses the published attribution equation. EAP-IG holds the activation difference fixed and averages gradients along an interpolation path in gate space, rather than the embedding space used by \citet{hanna2024}. EAP-GP applies the published GradPath updates in gate-coefficient space. No reference implementation of EAP-GP was released, so we follow its equations.

We check the shared ranking pipeline on a 325-edge GPT-2/IOI block graph with 16 examples. EAP-IG scores correlate with exact single-edge activation-patching effects at Spearman $0.545$; a random ranking scores $-0.055$. This tests whether our pipeline tracks edge effects. It does not establish numerical equivalence to the authors' implementations.

Each \texttt{knee.json} records the full-circuit baseline, pruning rounds, selected endpoint, and sealed-test score. Frozen baselines are stored as curves over edge budgets. We report seeds 11, 22, and 33. An earlier pilot seed is excluded because its capability probe overlapped with its test data.

\subsection{Selection and Evaluation Data}\label{app:matchprotocol}

Table~\ref{tab:matched-eapig-summary} breaks the headline reduction into four tasks. Each row includes all 24 model--seed runs.

\begin{table}[H]
 \centering
 \small
\caption{\textbf{Circuit-size reduction relative to frozen EAP-IG, by task.} Each task has 24 runs (eight models, three seeds). We select the smallest circuit within $0.05$ of each arm's own full-graph validation accuracy (\S\ref{sec:setup}). Reductions are geometric means of per-run ratios; the maximum is the largest single-run reduction.}
 \label{tab:matched-eapig-summary}
 \begin{tabular}{lrrrrr}
 \toprule
 Task & Runs & Ratios available & Reduction (geom.\ mean) & $C_1$ fewer edges & Max reduction \\
 \midrule
 IOI & 24 & 24 & $17.7{\times}$ & 24/24 & $209.2{\times}$ \\
 Agreement & 24 & 24 & $5.1{\times}$ & 21/24 & $30.6{\times}$ \\
 Induction & 24 & 24 & $18.2{\times}$ & 24/24 & $316.0{\times}$ \\
 Docstring & 24 & 24 & $2.7{\times}$ & 19/24 & $35.5{\times}$ \\
 \midrule
 Overall & 96 & 96 & $8.1{\times}$ & \textbf{88/96} & $316.0{\times}$ \\
 \bottomrule
 \end{tabular}
\end{table}

Table~\ref{tab:matchprotocol} records which data are used to rank edges, select circuits, and report each quantity.

Each arm is evaluated against its own full graph. This could favor $C_1$ if adaptation lowered full-graph accuracy. The adapted-minus-original difference is \AdaptedFullDelta{} at the validation median, although the adapted graph scores lower in 42 of \NRuns{} runs and higher in 24. Using the \emph{original} graph's accuracy minus $0.05$ as an absolute bar, all \AbsBarVal{} $C_1$ endpoints still qualify on validation and \AbsBarTest{} qualify on sealed test. Thus, the arm-relative rule does not account for the reported reduction.

\begin{table}[htbp]
\centering
\small
\setlength{\tabcolsep}{3pt}
\caption{\textbf{Data used for selection and reporting.} One rule determines the primary sizes. The analyses of \S\ref{sec:e4} re-evaluate those circuits on the sealed test split, using a subsample when they require thousands of masked circuits. We verify each sample size from the stored accuracy denominator rather than the configuration.}
\label{tab:matchprotocol}
\begin{tabular}{@{}
  >{\raggedright\arraybackslash}p{0.22\linewidth}
  >{\raggedright\arraybackslash}p{0.10\linewidth}
  >{\raggedright\arraybackslash}p{0.27\linewidth}
  >{\raggedright\arraybackslash}p{0.19\linewidth}
  >{\raggedright\arraybackslash}p{0.11\linewidth}@{}}
\toprule
Quantity & Ranked on & Selected on & Reported on & Examples \\
\midrule
Primary sizes, every arm & train & validation (own full graph $-0.05$) & sealed test & 2000 / 1000 \\
Matched frozen circuit (\S\ref{sec:e4}) & train & matched to $C_1$ test endpoint & sealed test & 1000 \\
Subset enumeration & -- & circuits fixed above & sealed-test subsample & 200 \\
Single-edge ablations & -- & circuits fixed above & sealed-test subsample & 96 \\
Whole-set ablation, pairs, roles, specificity & -- & circuits fixed above & sealed-test subsample & 200 \\
\bottomrule
\end{tabular}
\end{table}

Frozen EAP and EAP-IG curves and $C_1$ endpoints are scored on the same 1000 sealed-test examples. When an analysis matches a frozen circuit to $C_1$'s test accuracy, it uses that full split. Edge rankings still use training data only.

\subsection{Per-Task Answer Sets}\label{app:answersets}

All four tasks use restricted-choice accuracy at one answer position, requiring the correct token to outrank a task-defined set of alternatives. IOI and agreement use two-way minimal pairs: the indirect object against the subject, and a verb against the same verb in the wrong number. Induction selects from a fixed pool of plausible copy targets filtered to valid tokens for each model, with the reciprocal of the pool size as its chance baseline. Docstring completion compares the four single-token argument names in the signature, with the correct third argument required to outrank the others.

\section{Headline Statistics and Per-Task Reductions}\label{app:stats}

The size comparison uses 96 runs, but the three seeds for a task--model pair are related. The statistics below account for that dependence.

Seeds for one task--model pair share a dataset, base model, and graph. We therefore summarize each of the 32 pairs by its seed median. For reduction ratios, we report the geometric mean \citep{fleming1986}. The arithmetic mean is $29.6\times$ with interval $[11.7, 52.6]$; only 17 of \NRuns{} runs reach that mean. The geometric mean is less driven by those large ratios and agrees with the per-task summary because each task contributes equally many runs.

Confidence intervals use a cluster bootstrap over task--model pairs with 20{,}000 resamples. Every run has a selected circuit under the rule in \S\ref{sec:setup}, so all \NCells{} pairs enter the test (Table~\ref{tab:matched-eapig-summary}). With eight pairs per task, the smallest attainable sign-test value is $p=0.0078$. We correct the five comparisons in Table~\ref{tab:arms} with Holm--Bonferroni; all remain significant at $0.05$, with largest adjusted $p=\HolmMaxP$. The task-level intervals are $[8.0, 32.9]$ for IOI, $[2.5, 9.7]$ for agreement, $[6.3, 61.1]$ for induction, and $[1.1, 6.7]$ for docstring. Treating all 96 runs as independent would give $p\approx10^{-29}$ and overstate the evidence.

For the matched frozen circuit in \S\ref{sec:e4}, we match $C_1$'s sealed-test accuracy. Its validation score was checked repeatedly during healing and could be optimistic out of sample.

Like every arm, $C_0$ is selected as its smallest circuit within $0.05$ of its own full-graph validation accuracy. It stops within a median $0.03$ validation accuracy of $C_1$. The per-task $C_1$--$C_0$ gap is \CzeroTaskIOI{} on IOI, \CzeroTaskAgr{} on agreement, \CzeroTaskInd{} on induction, and \CzeroTaskDoc{} on docstring.

The overall \ConeVsEapig{} reduction might be driven by cells where the capability gate rarely binds. Excluding Gemma leaves \ShrinkNoGemma{}, with $C_1$ smaller in \ShrinkNoGemmaCells{} of 24 cells. Restricting to IOI and induction, whose own accuracy sets the endpoint, gives \ShrinkAccBound{} over 16 of 16 cells. Applying both restrictions gives \ShrinkBothRestr{} over 12 of 12 (\S\ref{sec:limitations}).

We also apply the common EAP-IG sweep to each task-SFT model. SFT alone concentrates the circuit: its selected circuits are \SftVsFrozen{} of the unmodified model's and smaller in 24 of \SftCells{} cells. $C_1$ is still a geometric-mean \SftVsCone{} smaller than SFT followed by discovery and wins in \SftVsConeWins{} of \SftCells{} cells. SFT wins only on agreement and docstring cells with weaker endpoint accuracy (Table~\ref{tab:endpoints}).

On sealed test, \SealedPass{} of \NRuns{} $C_1$ endpoints meet the tolerance. Both $C_1$ and EAP-IG meet their own tolerances in \SealedBoth{} runs; $C_1$ is smaller in \SealedBothWins{} of those.

The run logs make validation reuse measurable. The controller makes a median of 32 validation decisions per run, counting accepted rounds, rejected proposals, aborted heals, and re-heal probes (interquartile range 25.25--37, maximum 52). The sealed test split is read once per endpoint and never enters those decisions.

\subsection{Fixed-Performance Discovery Payoff}\label{app:budget}

We repeat circuit selection at $90\%$ of each arm's full-graph accuracy instead of the main $0.05$ tolerance (Table~\ref{tab:budget-payoff}).

\begin{table}[H]
\centering
\small
\caption{\textbf{Edges needed to recover $90\%$ of full-model accuracy} (32 cells, three seeds). We compare $C_1$ with the smallest frozen alternative among EAP, EAP-IG, random ranking, ACDC, and $C_0$. ACDC covers \AcdcCells{} cells on the two smallest graphs, where runs take $3.4$--$5.9$ hours; we do not run it on larger graphs. It uses the attribution arms' answer-slot KL metric. $C_0$ is best in 78 of 96 runs and ACDC in one. Ratios above $1\times$ favor condensation. $C_1$ loses eight runs, all on agreement or docstring, where the capability gate stops it early (\S\ref{sec:limitations}).}
\label{tab:budget-payoff}
\begin{tabular}{lrrrr}
\toprule
Task & Runs & Median reduction & Wins & $\geq 3\times$ \\
\midrule
IOI & 24 & $14.2{\times}$ & 24/24 & 21/24 \\
Agreement & 24 & $3.8{\times}$ & 21/24 & 16/24 \\
Induction & 24 & $8.7{\times}$ & 24/24 & 21/24 \\
Docstring & 24 & $1.7{\times}$ & 19/24 & 8/24 \\
\midrule
Overall & 96 & $5.2{\times}$ & 88/96 & 66/96 \\
\bottomrule
\end{tabular}
\end{table}

\section{Endpoints, Faithfulness, and Capability Retention}\label{app:klgrid}

Table~\ref{tab:klgrid} reports sealed-test KL from each circuit to its adapted full model, alongside the tolerance pass counts against the original model's accuracy. Five of 96 endpoints miss the tolerance, by a median \SealedShortfall{}. Agreement has the highest KL and the lowest original-model accuracy ($0.764$ at the test median).

\begin{table}[htbp]
\centering
\small
\setlength{\tabcolsep}{5pt}
\caption{\textbf{Sealed-test faithfulness across the head grid} (96 runs, medians with interquartile ranges). Agreement is the only task where circuit accuracy exceeds original-model accuracy at the median ($0.777$ versus $0.764$). Interchange ablation can sharpen the margin by removing distractor pathways outside the circuit, which also explains logit-difference recoveries above $1$.}
\label{tab:klgrid}
\begin{tabular}{lccccc}
\toprule
Task & Median KL & KL IQR & Median circuit acc. & Median original acc. & Sealed pass \\
\midrule
IOI & 0.048 & [0.045,\,0.063] & 0.978 & 0.996 & 24/24 \\
Induction & 0.081 & [0.060,\,0.124] & 0.951 & 0.965 & 19/24 \\
Docstring & 0.089 & [0.074,\,0.138] & 0.994 & 1.000 & 24/24 \\
Agreement & 0.254 & [0.145,\,0.343] & 0.777 & 0.764 & 24/24 \\
\midrule
All & 0.088 & -- & -- & -- & 91/96 \\
\bottomrule
\end{tabular}
\end{table}

\subsection{General-Capability Retention}\label{app:capability}

With every gate open, the adapted models have a median $9\%$ higher WikiText-2 perplexity \citep{merity2017wikitext} than the originals; the worst ratio is $1.7\times$. LAMBADA accuracy \citep{paperno2016lambada} falls by $5$--$7$ points (Table~\ref{tab:capability}). The controller's $1.05$ gate uses a separate mixed-domain probe. If that threshold were applied to WikiText-2, $18$--$19$ of $32$ cells would exceed it.

\begin{table}[H]
\centering
\footnotesize
\setlength{\tabcolsep}{4pt}
\caption{\textbf{General-capability retention of $C_1$.} We report the WikiText-2 perplexity ratio (adapted/original), per-token KL from the original, and change in LAMBADA last-token accuracy. The mixed-domain gate corpus is separate from both datasets. Values are medians across 32 cells for each seed. Seed 22 has 31 cells, giving 95 of \NRuns{} runs in total.}
\label{tab:capability}
\begin{tabular}{lccccc}
\toprule
Seed & Median ratio & Worst ratio & Median KL & Median $\Delta$LAMBADA & Cells $>1.05$ \\
\midrule
11 & 1.088 & 1.672 & 0.278 & $-0.068$ & 19/32 \\
22 & 1.091 & 1.742 & 0.255 & $-0.057$ & 19/31 \\
33 & 1.087 & 1.621 & 0.262 & $-0.052$ & 18/32 \\
\bottomrule
\end{tabular}
\end{table}

\section{Control Protocols and Per-Cell Results}\label{app:controls}

The controls change one part of the procedure at a time (Table~\ref{tab:controls-summary}). Tables~\ref{tab:b1} and~\ref{tab:b4} give the healing-only and head-swap results by cell.

\paragraph{Healing alone does not concentrate.} We keep every edge open and run healing with the same capability gate. The EAP-IG participation ratio measures how many edges share the attribution mass. Its median end-to-start ratio is $0.99$; 92 of 96 runs stay within $15\%$ of their starting value. The four exceptions are IOI runs on Qwen3 models, with the lowest ratio at $0.76$.

\paragraph{Attribution ranking is necessary.} We replace EAP-IG with random ordering or cut the highest-ranked edges first. Random ordering returns circuits a median \BtwoRandom{} larger than $C_1$ across \NCells{} cells and \NSeeds{} seeds; even its closest run is \BtwoRandomMin{} larger (Table~\ref{tab:b2}). Reverse ordering, run once per cell, accepts no cuts and retains the full graph.

\begin{table}[htbp]
\centering
\small
\caption{\textbf{Ranking controls.} Only the edge order changes. Values are seed medians except for deterministic anti-ranking, which retains the full graph. Random and anti-ranked accuracies are measured on validation, while EAP-IG accuracy is measured on the sealed test split. These controls do not use a $C_1$ endpoint.}
\label{tab:b2}
\begin{tabular}{llccc}
\toprule
Task & Model & EAP-IG (ours) & Random & Anti-ranked \\
& & edges @ acc & edges @ acc & edges @ acc \\
\midrule
IOI & GPT-2 & \textbf{58} @ 0.94 & 1484 @ 0.94 & 2041 @ 0.98 (full) \\
Agreement & Llama-3.2-1B & \textbf{263} @ 0.71 & 7644 @ 0.60 & 8993 @ 0.70 (full) \\
Induction & Gemma-3-1B & \textbf{3} @ 0.90 & 1197 @ 0.92 & 3537 @ 0.89 (full) \\
Docstring & Llama-3.2-1B & \textbf{519} @ 1.00 & 7128 @ 1.00 & 8993 @ 1.00 (full) \\
\bottomrule
\end{tabular}
\end{table}

\paragraph{Target specificity across all four behaviors.} After condensing one target, we run frozen discovery for every behavior in both the adapted and original models, using the same three seeds. The targeted behavior becomes \TargetShrinkInduction{} easier to localize for induction, \TargetShrinkAgreement{} for agreement, and \TargetShrinkDocstring{} for docstring, alongside the IOI result in \S\ref{sec:controls}. The 12 off-target direction medians stay at \OffTargetShrink{} across 288 measurements. Individual cells do move: 58 measurements exceed $1\times$, mostly on agreement, where the non-monotone accuracy curve makes budget selection less stable (\S\ref{sec:limitations}).

\paragraph{The circuit is load-bearing and head-specific.} Closing every edge leaves the adapter at chance, so it cannot carry the task alone. To test the selected heads, we re-point each retained head$\to$logits edge to another head in the same layer. Accuracy falls to chance or below in \SwapChance{} of \NCells{} circuits (Table~\ref{tab:b4}). Adding a random edge leaves accuracy unchanged, which ties the drop to the head substitutions.

The all-edges-off reference can fall below guessing because every edge then reads a corrupted activation. Against that stricter reference, \SwapFloorCount{} of \NCells{} swapped circuits finish within $0.13$. The two above-chance cases are IOI/Llama-3.2-1B (\SwapLlamaAcc{}) and agreement/Gemma-3-1B (\SwapGemmaAcc{}), both well below intact accuracy; these are two-choice tasks with chance level $0.50$.

\subsection{Healing-Only Control}\label{app:b1}

Table~\ref{tab:b1} shows representative healing-only runs; the grid-wide median ratio is \BoneParticipation{} (\S\ref{sec:controls}).

\begin{table}[H]
\centering
\small
\caption{\textbf{Healing-only control at head granularity.} Representative cells are shown, with EAP-IG participation ratios averaged over seeds 11/22/33. Higher PR means attribution is spread across more edges. The grid-wide median ratio is $0.99$ across 32 cells. Runs start from fresh checkpoints rather than a $C_1$ endpoint.}
\label{tab:b1}
\begin{tabular}{llccc}
\toprule
Task & Model & PR before & PR after (no-prune heal) & Ratio \\
\midrule
IOI & GPT-2 & 50.7 & 50.5 & 1.00 \\
IOI & Llama-3.2-1B & 331.0 & 328.3 & 0.99 \\
IOI & Qwen3-4B & 1017.7 & 876.1 & 0.86 \\
Agreement & Llama-3.2-1B & 267.9 & 261.7 & 0.98 \\
Agreement & Llama-3.2-3B & 479.3 & 475.2 & 0.99 \\
Induction & Gemma-3-1B & 47.1 & 46.7 & 0.99 \\
Docstring & Qwen3-1.7B & 97.5 & 97.0 & 0.99 \\
\bottomrule
\end{tabular}
\end{table}

\subsection{Conduit and Head-Specificity Control}\label{app:b4}

Table~\ref{tab:b4} lists the head-swap result for each cell (\S\ref{sec:controls}).

\begin{table}[H]
\centering
\small
\caption{\textbf{Sibling-swap control} (head granularity, all \NCells{} cells). Values are three-seed medians on 200 examples. \emph{Swapped} counts re-pointed head$\to$logits edges. \emph{All off} and \emph{Kept} give accuracy with every edge closed and with the circuit intact. Chance is $0.50$ for the two-choice tasks, $0.25$ for docstring, and the reciprocal of the induction pool size. \emph{To chance} marks swapped circuits no better than guessing.}
\label{tab:b4}
\setlength{\tabcolsep}{3pt}%
\begin{tabular}{llrrrrrrc}
\toprule
Task & Model & Edges & Swapped & Kept & All off & Chance & Swapped acc. & To chance \\
\midrule
IOI & Gemma-3-4B & 4 & 2 & 1.000 & 0.555 & 0.50 & 0.550 & yes \\
 & Llama-3.2-3B & 10 & 3 & 0.985 & 0.510 & 0.50 & 0.490 & yes \\
 & Gemma-3-1B & 12 & 5 & 0.985 & 0.510 & 0.50 & 0.435 & yes \\
 & Llama-3.2-1B & 18 & 7 & 0.980 & 0.515 & 0.50 & 0.580 & no \\
 & GPT-2 & 58 & 17 & 0.970 & 0.485 & 0.50 & 0.475 & yes \\
 & Qwen3-1.7B & 94 & 23 & 0.975 & 0.510 & 0.50 & 0.560 & yes \\
 & Qwen3-4B & 135 & 34 & 0.965 & 0.525 & 0.50 & 0.495 & yes \\
 & Qwen3-0.6B & 169 & 34 & 0.975 & 0.520 & 0.50 & 0.515 & yes \\
Agreement & Gemma-3-1B & 125 & 4 & 0.815 & 0.225 & 0.50 & 0.605 & no \\
 & Gemma-3-4B & 180 & 13 & 0.710 & 0.375 & 0.50 & 0.390 & yes \\
 & Qwen3-4B & 190 & 18 & 0.735 & 0.190 & 0.50 & 0.400 & yes \\
 & Qwen3-1.7B & 205 & 13 & 0.805 & 0.255 & 0.50 & 0.235 & yes \\
 & Llama-3.2-1B & 263 & 31 & 0.700 & 0.290 & 0.50 & 0.525 & yes \\
 & Llama-3.2-3B & 314 & 22 & 0.695 & 0.280 & 0.50 & 0.515 & yes \\
 & GPT-2 & 453 & 72 & 0.815 & 0.170 & 0.50 & 0.450 & yes \\
 & Qwen3-0.6B & 1054 & 138 & 0.850 & 0.150 & 0.50 & 0.180 & yes \\
Induction & Gemma-3-1B & 3 & 1 & 0.925 & 0.000 & 0.05 & 0.000 & yes \\
 & Gemma-3-4B & 4 & 2 & 0.925 & 0.000 & 0.05 & 0.000 & yes \\
 & Qwen3-4B & 88 & 33 & 0.985 & 0.000 & 0.05 & 0.000 & yes \\
 & Llama-3.2-3B & 91 & 25 & 0.985 & 0.000 & 0.05 & 0.000 & yes \\
 & Llama-3.2-1B & 104 & 38 & 0.970 & 0.000 & 0.05 & 0.000 & yes \\
 & GPT-2 & 160 & 50 & 0.930 & 0.000 & 0.05 & 0.000 & yes \\
 & Qwen3-1.7B & 228 & 57 & 1.000 & 0.000 & 0.05 & 0.000 & yes \\
 & Qwen3-0.6B & 404 & 76 & 0.980 & 0.000 & 0.05 & 0.000 & yes \\
Docstring & Gemma-3-1B & 3 & 1 & 0.970 & 0.000 & 0.25 & 0.000 & yes \\
 & Gemma-3-4B & 7 & 3 & 0.985 & 0.000 & 0.25 & 0.000 & yes \\
 & GPT-2 & 144 & 34 & 0.415 & 0.000 & 0.25 & 0.000 & yes \\
 & Qwen3-1.7B & 475 & 80 & 0.995 & 0.000 & 0.25 & 0.000 & yes \\
 & Llama-3.2-1B & 519 & 111 & 1.000 & 0.000 & 0.25 & 0.000 & yes \\
 & Qwen3-4B & 608 & 121 & 1.000 & 0.000 & 0.25 & 0.000 & yes \\
 & Llama-3.2-3B & 697 & 119 & 1.000 & 0.000 & 0.25 & 0.000 & yes \\
 & Qwen3-0.6B & 1877 & 157 & 1.000 & 0.000 & 0.25 & 0.000 & yes \\
\bottomrule
\end{tabular}

\end{table}

\subsection{Joint Gate and Adapter Training}\label{app:joint}

The joint-training arm removes the prune-and-heal loop. It trains gates and the LoRA adapter together for 3000 steps under an $L_0$ sparsity penalty, using the same head-level graph, splits, restricted-choice metrics, and binary edge masks. Base-model weights stay frozen; the LoRA parameters train. We record the norm of the LoRA $B$ matrices, initialized at zero, to confirm that the adapter changed.

We choose the smallest admissible configuration separately for each cell. It must have at least one edge, clear the \emph{original} model's validation accuracy minus $0.05$, and stay within the controller's $1.05$ capability-probe ratio. We use the original-model bar because joint training can lower its own gates-open accuracy; a bar relative to that damaged model would be easier to clear. Sealed-test accuracy is reported after selection, not used to choose the configuration (Table~\ref{tab:w3joint}).

Four pilot cells score 16 configurations each, covering two preservation weights and eight sparsity weights. The other 28 cells score five sparsity weights at the capability-preserving preservation weight; the remaining planned configurations were not run. This limited sweep may miss better joint-training circuits. Per-cell tuning favors this arm relative to $C_1$'s fixed controller settings. The observed \WThreeSmaller{} of \WThreeCells{} wins describe the configurations tested here, not a bound on what a wider search could find.

Table~\ref{tab:w3joint} gives each cell's result and the number of admissible configurations among those scored. A zero means none of the tested configurations passed all three criteria; it does not rule out untested settings.

\begin{table}[H]
\centering\small
\setlength{\tabcolsep}{5pt}
\caption{\textbf{Joint gates and adapter across all \WThreeCells{} cells.} $C_1$/W3 above $1$ means the baseline returns fewer edges; below $1$, the controller does. Dashes mean no tested configuration passes the accuracy and capability criteria. The final column gives admissible configurations out of those scored.}
\label{tab:w3joint}
\begin{tabular}{llrrrl}
\toprule
Task & Model & W3 edges & $C_1$ edges & $C_1$/W3 & Admissible configs \\
\midrule
IOI & GPT-2 & 5 & 58 & 11.6$\times$ & 2 of 16 \\
 & Llama-3.2-1B & 22 & 18 & 0.8$\times$ & 2 of 5 \\
 & Llama-3.2-3B & none & 10 & --- & 0 of 5 \\
 & Gemma-3-1B & 3 & 12 & 4.0$\times$ & 3 of 5 \\
 & Gemma-3-4B & 3 & 4 & 1.3$\times$ & 4 of 5 \\
 & Qwen3-0.6B & none & 169 & --- & 0 of 5 \\
 & Qwen3-1.7B & none & 94 & --- & 0 of 5 \\
 & Qwen3-4B & none & 135 & --- & 0 of 5 \\
Agreement & GPT-2 & none & 453 & --- & 0 of 5 \\
 & Llama-3.2-1B & 21 & 263 & 12.5$\times$ & 2 of 16 \\
 & Llama-3.2-3B & none & 314 & --- & 0 of 5 \\
 & Gemma-3-1B & 674 & 125 & 0.2$\times$ & 1 of 5 \\
 & Gemma-3-4B & none & 180 & --- & 0 of 5 \\
 & Qwen3-0.6B & none & 1054 & --- & 0 of 5 \\
 & Qwen3-1.7B & 104 & 205 & 2.0$\times$ & 3 of 5 \\
 & Qwen3-4B & 633 & 190 & 0.3$\times$ & 1 of 5 \\
Induction & GPT-2 & none & 160 & --- & 0 of 5 \\
 & Llama-3.2-1B & none & 104 & --- & 0 of 5 \\
 & Llama-3.2-3B & none & 91 & --- & 0 of 5 \\
 & Gemma-3-1B & 213 & 3 & 0.0$\times$ & 2 of 16 \\
 & Gemma-3-4B & 167 & 4 & 0.0$\times$ & 2 of 5 \\
 & Qwen3-0.6B & none & 404 & --- & 0 of 5 \\
 & Qwen3-1.7B & none & 228 & --- & 0 of 5 \\
 & Qwen3-4B & none & 88 & --- & 0 of 5 \\
Docstring & GPT-2 & none & 144 & --- & 0 of 5 \\
 & Llama-3.2-1B & none & 519 & --- & 0 of 16 \\
 & Llama-3.2-3B & none & 697 & --- & 0 of 5 \\
 & Gemma-3-1B & 128 & 3 & 0.0$\times$ & 3 of 5 \\
 & Gemma-3-4B & 47 & 7 & 0.1$\times$ & 3 of 5 \\
 & Qwen3-0.6B & none & 1877 & --- & 0 of 5 \\
 & Qwen3-1.7B & none & 475 & --- & 0 of 5 \\
 & Qwen3-4B & none & 608 & --- & 0 of 5 \\
\bottomrule
\end{tabular}

\end{table}

\subsection{Seed Stability Detail}\label{app:seedstab}

Table~\ref{tab:seedstab} compares circuit edges across seeds. Edge Jaccard divides the overlap by the union; containment asks what share of the smaller circuit lies in the larger. The null draws two random subsets of the observed sizes from the full graph. On checked cells, 2,000 simulated draws per pair agree with the analytic null within $0.001$.

\begin{table}[H]
\centering
\small
\caption{\textbf{Seed stability of $C_1$ across all 32 cells} (head granularity, seeds 11/22/33). Values are per-task medians over seed pairs. $\times$null is the observed edge Jaccard divided by its expectation for independent random subsets of the same sizes. High containment and above-null Jaccard indicate consistent, often nested circuits across seeds.}
\label{tab:seedstab}
\begin{tabular}{lccc}
\toprule
Task & Containment & Edge Jaccard & $\times$ null \\
\midrule
IOI & 0.93 & 0.53 & $205\times$ \\
Agreement & 0.89 & 0.68 & $59\times$ \\
Induction & 0.95 & 0.83 & $243\times$ \\
Docstring & 0.94 & 0.76 & $45\times$ \\
\bottomrule
\end{tabular}
\end{table}

\section{Verification Detail and Analysis Cost}\label{app:payoff-details}

Table~\ref{tab:analysis-cost} times exact single-edge patching on IOI/GPT-2. It measures the saving from testing fewer edges after condensation and excludes the cost of producing the circuit.

\begin{table}[H]
\centering
\small
\caption{\textbf{Exact-patching cost on IOI/GPT-2.} Forward-pass and wall-clock reductions after restricting candidates to the condensed graph.}
\label{tab:analysis-cost}
\begin{tabular}{lrrrrr}
\toprule
Candidate set & Condensed & Original & Passes & Time & Speedup \\
\midrule
Head edges & 70 & 256 & 1280$\to$350 & 137.02s$\to$37.66s & $3.6{\times}$ \\[2pt]
\multicolumn{6}{@{}p{.95\linewidth}@{}}{\footnotesize Measured on an earlier 70-edge IOI/GPT-2 endpoint; the reported endpoint is 58 edges (Table~\ref{tab:endpoints}), so this understates the saving.} \\
\bottomrule
\end{tabular}
\end{table}

\subsection{Total Compute and Break-Even}\label{app:cost}

Each of the 96 head-level runs uses one 40GB or 80GB NVIDIA A100. Median runtime is $2.95$ hours, ranging from $0.33$ hours on GPT-2 to $17.5$ on Qwen3-4B. Healing takes about $11\%$ of the time; repeated EAP-IG ranking and gated evaluation take most of the rest. Exact single-edge patching on the condensed graph is \AnalysisSpeedup{} faster per analysis (Table~\ref{tab:analysis-cost}), so many repeated analyses are needed to recover the training cost.

The larger gain is in analyses that scale faster than linearly with edge count. Exhaustive subset and pair tests would require a median \PairFrozenCostMedian{} more ablations on matched frozen circuits (interquartile range \PairFrozenCostIQR{}). Only \PairRunCells{} of 32 frozen cells have a projected all-pairs runtime under one day (\S\ref{sec:e4}).

\section{Legibility and Original-Model Checks}\label{app:ioi-roles}

We compare the condensed IOI circuit with the head roles mapped by \citet{wang2023}. Across seeds 11/22/33, it contains a median \RoleCondHeads{} heads ($21$--$27$), including \RoleCondRoles{} with published roles ($17$--$19$). It covers every stage of the published algorithm except the previous-token stage. The matched-accuracy frozen circuit contains \RoleFrozenHeads{} heads ($61$--$82$), with \RoleFrozenRoles{} named in all three seeds. The condensed circuit has a higher share of named heads but omits \RoleDropped{} of the \WangHeads{} published heads, including three of eight backup name movers. The frozen circuit retains all but one of those eight.

\subsection{Interchange Intervention Detail}\label{app:iia}

Role counts identify heads by name. Interchange interventions \citep{geiger2021,geiger2022} ask whether changing the variable assigned to those heads changes the answer as Wang et al.'s algorithm predicts (Table~\ref{tab:iia}).

\begin{table}[H]
\centering
\footnotesize
\setlength{\tabcolsep}{3pt}
\caption{\textbf{Interchange intervention accuracy under Wang et al.'s IOI algorithm} (GPT-2, three seeds, 200 prompt pairs each). Every arm uses the same hypothesis and edge indices. We assign the output variable to named name movers and the intermediate variable to subject-inhibition heads feeding them. Higher accuracy means the intervention more often produces the predicted answer. ``Random'' uses unnamed heads; ``Sham'' tests a task without a published role map. The coarse all-in-circuit variant saturates and serves only as a control.}
\label{tab:iia}
\begin{tabular}{lccc}
\toprule
Aligned variable & $C_1$ condensed & Frozen (matched acc.) & Unmodified model \\
\midrule
Output (name-mover $\rightarrow$ logits) & 0.995 & 0.985 & 0.995 \\
Intermediate (subject inhibition) & \textbf{0.930} & 0.735 & 0.645 \\
\quad per-seed range & $0.910$--$0.945$ & $0.715$--$0.740$ & $0.640$--$0.650$ \\
\midrule
Random-edge control (intermediate) & 0.03 & 0.04 & 0.01 \\
Sham hypothesis (different task) & 0.00 & 0.01 & 0.01 \\
\bottomrule
\end{tabular}
\end{table}

Patching the named name-mover outputs scarcely separates the arms: the condensed circuit, frozen circuit, and unmodified model score $0.995$, $0.985$, and $0.995$. The subject-inhibition intervention is more discriminating (Table~\ref{tab:iia}). Circuit size may still affect the comparison. Cutting the frozen circuit to $C_1$'s edge count leaves it at chance in two of three seeds. In the third, it remains functional at $0.885$ task accuracy, and $C_1$ scores $0.965$ versus its $0.845$ on the intervention.

The next three checks compare the circuit with the original model. They come from an earlier, coarser replication that treats an attention or MLP block as one component. The main results use head-level graphs, so these block-level values do not enter the headline statistics.

\subsection{Block-Level Endpoint Reproduction}\label{app:repro}

We reran 23 block-level cells from scratch with the same controller and seed. The median change in smallest-faithful edge count is zero; most endpoints reproduce exactly. One IOI/Qwen3-4B rerun moves from 18 to 67 edges at matched accuracy, showing that the stopping point can vary. This check is separate from the head-level, three-seed comparison in Appendix~\ref{app:seedstab}.

\subsection{Block-Level Logit-Difference Recovery}\label{app:ldrecovery}

For comparison with EAP-style evaluations, Table~\ref{tab:origfaith} reports normalized logit-difference recovery for the 32 block-level cells and the head-level IOI anchor: $(m(\text{circuit})-m(\text{ablated}))/(m(\text{full})-m(\text{ablated}))$. Here $m$ is the contrastive logit margin for IOI and agreement. The earlier block-level induction and docstring runs use the answer logit because their restricted pools contain more than one foil. Those values do not enter the current head-level headline. A value of $1$ reproduces the full model's margin, while values above $1$ mean interchange ablation sharpens it. Median recovery is $0.95$ (range $0.53$--$1.25$) on the sealed test split without retraining.

\subsection{Original-Model Faithfulness}\label{app:origfaith}

Because condensation changes weights, a circuit can fit the adapted model without reproducing the original model. The healing objective encourages agreement through $\kl(\text{orig}\,\|\,\text{masked})$, and the following checks measure how close the two remain.

\paragraph{The current head-level grid.} Across all \NRuns{} current head-level runs, the circuit selects the same answer as the unmodified model on a median \OrigAgree{} of sealed-test examples, with a median token-level KL of \OrigKlGrid{}. These are the grid-wide values reported in \S\ref{sec:e5}. On the IOI/GPT-2 anchor, the per-seed KL to the original model is $0.045$--$0.068$ (median \KlHeadCond{}).

\paragraph{The block-level replication.} Across the 32 block-level cells, the circuit matches the original model's task choice on a median $95.5\%$ of sealed-test examples, with a median token-level KL of $0.196$. Agreement has the lowest choice consistency at $0.74$--$0.95$. Docstring/GPT-2 is also weak because its base model scores only $0.59$. For comparison, the head-level IOI anchor has KL \KlHeadCond{} to the original GPT-2, against \KlHeadFrozen{} for its matched frozen circuit. Table~\ref{tab:origfaith} reports both sets of results. The block-level values are supporting checks rather than a grid-wide claim about the current head endpoints.

\begin{table}[H]
\centering\small
\setlength{\tabcolsep}{5pt}
\caption{\textbf{Original-model faithfulness and logit-difference recovery.} Results combine the block-level replication with the head-level IOI anchor. Match($M$) is the fraction of sealed-test examples where the circuit and original model select the same candidate. $\kl(M\|\hat{C})$ and $\kl(M'\|\hat{C})$ compare the circuit with the original and adapted full models. LD recovery is $(m(\text{circuit})-m(\text{ablated}))/(m(\text{full})-m(\text{ablated}))$, where $1$ reproduces the full model's margin. The head-level IOI row gives three-seed medians and has no LD value because that evaluation exists only at block granularity.}
\label{tab:origfaith}
\begin{tabular}{llrrrrrr}
\toprule
Task & Model & Edges & Acc. & Match($M$) & $\kl(M\|\hat{C})$ & $\kl(M'\|\hat{C})$ & LD rec. \\
\midrule
IOI & GPT-2 & 13 & 0.96 & 0.95 & 0.079 & 0.025 & 0.58 \\
 & Llama-3.2-1B & 15 & 0.98 & 0.99 & 0.217 & 0.047 & 0.95 \\
 & Llama-3.2-3B & 12 & 0.955 & 0.95 & 0.067 & 0.040 & 0.58 \\
 & Gemma-3-1B & 4 & 0.995 & 0.99 & 0.118 & 0.046 & 1.23 \\
 & Gemma-3-4B & 4 & 0.995 & 0.99 & 0.072 & 0.043 & 0.91 \\
 & Qwen3-0.6B & 71 & 0.98 & 0.96 & 0.157 & 0.030 & 1.03 \\
 & Qwen3-1.7B & 29 & 0.945 & 0.94 & 0.153 & 0.066 & 0.57 \\
 & Qwen3-4B & 67 & 0.95 & 0.96 & 0.089 & 0.033 & 0.81 \\
Agreement & GPT-2 & 227 & 0.82 & 0.95 & 0.050 & 0.053 & 0.98 \\
 & Llama-3.2-1B & 12 & 0.79 & 0.76 & 0.324 & 0.226 & 1.18 \\
 & Llama-3.2-3B & 37 & 0.64 & 0.78 & 0.340 & 0.255 & 0.88 \\
 & Gemma-3-1B & 39 & 0.75 & 0.79 & 0.533 & 0.369 & 0.90 \\
 & Gemma-3-4B & 82 & 0.555 & 0.74 & 0.361 & 0.202 & 0.65 \\
 & Qwen3-0.6B & 377 & 0.86 & 0.91 & 0.242 & 0.251 & 0.97 \\
 & Qwen3-1.7B & 116 & 0.79 & 0.80 & 0.327 & 0.218 & 1.08 \\
 & Qwen3-4B & 98 & 0.665 & 0.78 & 0.357 & 0.305 & 1.06 \\
Induction & GPT-2 & 65 & 0.965 & 0.96 & 0.169 & 0.097 & 0.86 \\
 & Llama-3.2-1B & 52 & 0.94 & 0.94 & 0.159 & 0.045 & 1.02 \\
 & Llama-3.2-3B & 53 & 0.97 & 0.96 & 0.185 & 0.064 & 1.17 \\
 & Gemma-3-1B & 4 & 0.93 & 0.84 & 0.250 & 0.126 & 0.79 \\
 & Gemma-3-4B & 4 & 0.95 & 0.93 & 0.359 & 0.113 & 0.98 \\
 & Qwen3-0.6B & 193 & 0.965 & 0.95 & 0.255 & 0.064 & 0.94 \\
 & Qwen3-1.7B & 161 & 0.99 & 0.99 & 0.291 & 0.048 & 0.98 \\
 & Qwen3-4B & 91 & 0.985 & 0.98 & 0.310 & 0.073 & 1.25 \\
Docstring & GPT-2 & 24 & 0.415 & 0.59 & 0.227 & 0.154 & 1.10 \\
 & Llama-3.2-1B & 50 & 1.0 & 1.00 & 0.165 & 0.081 & 0.92 \\
 & Llama-3.2-3B & 95 & 1.0 & 1.00 & 0.149 & 0.119 & 0.85 \\
 & Gemma-3-1B & 4 & 1.0 & 1.00 & 0.157 & 0.079 & 0.83 \\
 & Gemma-3-4B & 4 & 0.945 & 0.94 & 0.270 & 0.171 & 0.94 \\
 & Qwen3-0.6B & 278 & 0.99 & 0.99 & 0.133 & 0.095 & 1.00 \\
 & Qwen3-1.7B & 208 & 1.0 & 1.00 & 0.178 & 0.111 & 1.18 \\
 & Qwen3-4B & 114 & 0.99 & 0.99 & 0.207 & 0.172 & 1.10 \\
IOI (head) & GPT-2 & 58 & 0.950 & 0.965 & 0.056 & 0.025 & --- \\
\bottomrule
\end{tabular}
\end{table}

\section{Error Prediction from Circuit Margins}\label{app:failpred}

Circuit-shift monitoring has used head-level signals to detect distribution shift in vision transformers \citep{peng2026insideout}. Here we test whether a circuit's restricted-choice margin ranks errors made by the original model. At matched edge count, median AUROC is \FailAurocCond{} for $C_1$ and \FailAurocFrozen{} for the frozen circuit; $C_1$ leads in \FailWins{} of \FailCells{} cells (Table~\ref{tab:failpred}). IOI has too few errors ($0$--$2\%$), and docstring has three example-specific foils rather than one fixed contrast, so neither is scored.

At its full faithful size, the frozen circuit often scores higher but uses $1.3$--$500\times$ more edges. On induction/Llama-3.2-1B, it reaches $0.907$ versus $0.898$ for $C_1$, using 600 rather than 104 edges. Agreement/GPT-2 is the one matched-size exception, with the frozen circuit ahead by $0.002$.

\begin{table}[H]
\centering
\small
\caption{\textbf{Predicting original-model errors in 11 cells.} AUROC uses each circuit's restricted-choice margin to rank the original model's errors. Values are per-cell medians over three seeds. The $C_1$ and frozen columns use the same edge budget $k$; the final column uses the full frozen circuit. Agreement/Qwen3-0.6B ties at the shown precision, with $C_1$ ahead by less than $0.001$. The original model's own margin defines the error labels and would score perfectly. IOI and docstring lack enough suitable errors or a fixed contrast, respectively.}
\label{tab:failpred}
\begin{tabular}{lrrccl}
\toprule
Model & Err.\ rate & $k$ & $C_1$ (matched) & Frozen (matched) & Frozen (full) \\
\midrule
\multicolumn{6}{l}{\emph{Agreement}} \\
\quad GPT-2 & $19\%$ & 453 & 0.983 & \textbf{0.985} & 0.992 (600e) \\
\quad Llama-3.2-1B & $35\%$ & 263 & \textbf{0.943} & 0.927 & 0.965 (1400e) \\
\quad Llama-3.2-3B & $34\%$ & 314 & \textbf{0.936} & 0.635 & 1.000 (20329e) \\
\quad Gemma-3-1B & $22\%$ & 125 & \textbf{0.829} & 0.639 & 0.931 (600e) \\
\quad Gemma-3-4B & $37\%$ & 180 & \textbf{0.823} & 0.675 & 0.981 (2000e) \\
\quad Qwen3-0.6B & $21\%$ & 1054 & 0.931 & 0.931 & 1.000 (13833e) \\
\quad Qwen3-1.7B & $24\%$ & 205 & \textbf{0.877} & 0.621 & 1.000 (13833e) \\
\quad Qwen3-4B & $26\%$ & 190 & \textbf{0.894} & 0.645 & 1.000 (43993e) \\
\multicolumn{6}{l}{\emph{Induction}} \\
\quad Llama-3.2-1B & $5\%$ & 104 & \textbf{0.898} & 0.690 & 0.907 (600e) \\
\quad Gemma-3-1B & $13\%$ & 3 & \textbf{0.690} & 0.586 & 0.807 (600e) \\
\quad Gemma-3-4B & $5\%$ & 4 & \textbf{0.796} & 0.333 & 0.834 (2000e) \\
\midrule
\multicolumn{3}{l}{Median (11 cells)} & \textbf{0.894} & 0.645 & 0.981 \\
\bottomrule
\end{tabular}
\end{table}

The margin also supports selective prediction: keeping the most-confident $70\%$ of inputs raises accuracy from $0.781$ to $0.911$, versus $0.844$ for the matched frozen circuit.

\section{Additional Payoff Detail}\label{app:payoff-extra}

The remaining checks ask how much can be learned from the smaller graph at a fixed inspection budget and how its size and interactions vary across cells.

\begin{figure}[htbp]
 \centering
 \includegraphics[width=0.72\linewidth]{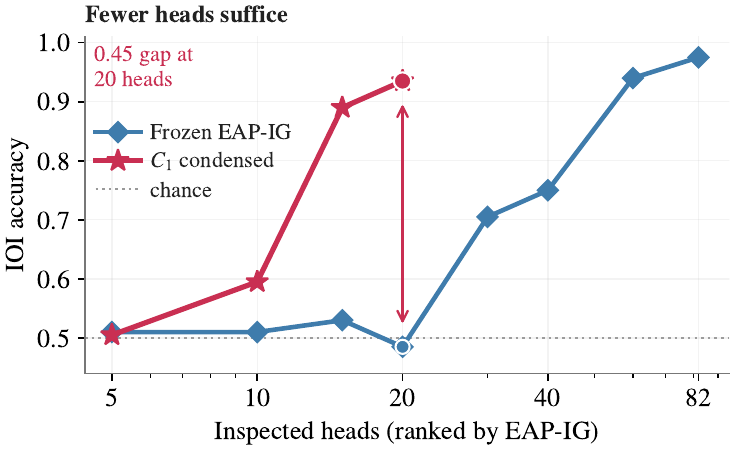}
 \caption{\textbf{Fixed-budget IOI legibility.} GPT-2 accuracy after retaining each arm's top EAP-IG-ranked heads, median over \NSeeds{} seeds.}
 \label{fig:legibility}

\end{figure}

\paragraph{A smaller set of heads carries the behavior.} With only the \HeadBudget{} highest-attribution IOI heads, the condensed circuit reaches \SuffCondensed{} accuracy. The frozen circuit remains at chance and needs \SuffFrozenBudget{} heads to match it (Figure~\ref{fig:legibility}). On induction, the same \SuffInductionBudget-head budget gives \SuffInductionCond{} for $C_1$ versus \SuffInductionFrozen{} for frozen discovery. Condensation does not increase the share of induction heads with identified roles; it makes fewer of them sufficient. The IOI interchange test is consistent with this result, although circuit size may affect that comparison (Appendix~\ref{app:iia}).

\begin{table}[H]
\centering
 \small
 \caption{\textbf{Condensed and frozen IOI endpoints} (GPT-2, matched accuracy \RoleMatchedAcc{}). Values are medians over seeds 11/22/33. Recall is the share of Wang's \WangHeads{} heads in a circuit, and precision is the share of its heads with a published role. For both arms, KL is $\kl(\text{orig}\,\|\,\text{circuit})$ against the unmodified model. Since entries are per-seed medians, the reported $F_1$ need not equal the value computed from median precision and recall. The frozen circuit is re-matched for each seed, giving its $256$--$400$ edge range.}
 \label{tab:roles}
\begin{tabular}{@{}lrr@{}}
 \toprule
 & $C_1$ & Frozen \\
 \midrule
 Edges & \textbf{58} & 256 \\
 Heads & \textbf{24} & 61 \\
 KL to original & \textbf{0.056} & 0.50 \\
 \midrule
 No published role & \textbf{7} & 36 \\
 Precision & \textbf{0.71} & 0.41 \\
 Recall & \RoleCondRecall & \textbf{\RoleFrozenRecall} \\
 $F_1$ & \textbf{\RoleCondFOne} & \RoleFrozenFOne \\
 \bottomrule
 \end{tabular}

\end{table}

\begin{figure}[htbp]
 \centering
 \includegraphics[width=0.75\linewidth]{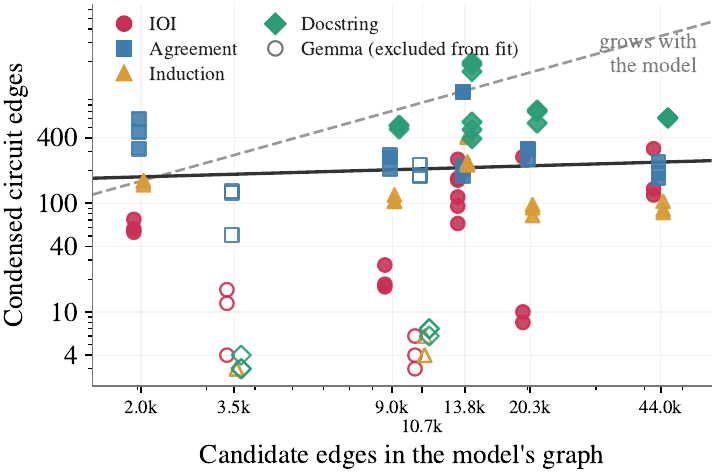}
 \caption{\textbf{Condensed circuit size against full graph size.} One point per run (93 total; low-accuracy docstring/GPT-2 is excluded). On log--log axes, the solid line fits the non-Gemma runs and the dashed line shows proportional growth. Hollow Gemma points are excluded from the fit because several endpoints have only 3--4 edges.}
 \label{fig:scaling}
\end{figure}

\paragraph{Circuit size does not track graph size in this grid.} Qwen3-4B has \GraphRatio{} times as many candidate edges as GPT-2, yet its condensed endpoints are of similar size (Figure~\ref{fig:scaling}). A descriptive log--log fit has slope \ScalingSlope{} ($95\%$ interval \ScalingCI{}), versus one for proportional growth. The pooled rank association is weak (Spearman \ScalingSpearman{}, $p=\ScalingSpearmanP$). We do not infer a scaling law from these eight models.

\begin{table}[htbp]
 \centering
 \scriptsize
 \setlength{\tabcolsep}{3pt}
 \caption{\textbf{Condensed endpoints at head granularity, all \NCells{} cells.} Median edges over seeds 11/22/33 with the seed range beneath, and median accuracy on the sealed 1000-example test split. ``Full'' is the complete head-level graph for that model.}
 \label{tab:endpoints}
 \begin{tabular}{lrrrrrrrrr}
 \toprule
 & & \multicolumn{2}{c}{IOI} & \multicolumn{2}{c}{Agreement} & \multicolumn{2}{c}{Induction} & \multicolumn{2}{c}{Docstring} \\
 \cmidrule(lr){3-4}\cmidrule(lr){5-6}\cmidrule(lr){7-8}\cmidrule(lr){9-10}
 Model & Full & Edges & Acc & Edges & Acc & Edges & Acc & Edges & Acc \\
 \midrule
 GPT-2 small & 2041 & 58\,{\scriptsize(54--71)} & 0.94 & 453\,{\scriptsize(317--595)} & 0.79 & 160\,{\scriptsize(148--160)} & 0.91 & 144\,{\scriptsize(118--175)} & 0.44 \\
 Llama-3.2-1B & 8993 & 18\,{\scriptsize(17--27)} & 0.97 & 263\,{\scriptsize(205--279)} & 0.71 & 104\,{\scriptsize(104--119)} & 0.97 & 519\,{\scriptsize(480--519)} & 1.00 \\
 Llama-3.2-3B & 20329 & 10\,{\scriptsize(8--267)} & 0.98 & 314\,{\scriptsize(254--316)} & 0.72 & 91\,{\scriptsize(77--97)} & 0.97 & 697\,{\scriptsize(545--720)} & 1.00 \\
 Gemma-3-1B & 3537 & 12\,{\scriptsize(4--16)} & 0.99 & 125\,{\scriptsize(51--129)} & 0.83 & 3\,{\scriptsize(3--3)} & 0.90 & 3\,{\scriptsize(3--4)} & 0.97 \\
 Gemma-3-4B & 10745 & 4\,{\scriptsize(3--6)} & 1.00 & 180\,{\scriptsize(176--226)} & 0.71 & 4\,{\scriptsize(4--6)} & 0.90 & 7\,{\scriptsize(6--7)} & 0.98 \\
 Qwen3-0.6B & 13833 & 169\,{\scriptsize(163--253)} & 0.98 & 1054\,{\scriptsize(1054--1054)} & 0.84 & 404\,{\scriptsize(404--417)} & 0.94 & 1877\,{\scriptsize(1628--1976)} & 1.00 \\
 Qwen3-1.7B & 13833 & 94\,{\scriptsize(65--114)} & 0.98 & 205\,{\scriptsize(178--221)} & 0.80 & 228\,{\scriptsize(226--240)} & 0.98 & 475\,{\scriptsize(391--559)} & 1.00 \\
 Qwen3-4B & 43993 & 135\,{\scriptsize(119--317)} & 0.96 & 190\,{\scriptsize(169--242)} & 0.76 & 88\,{\scriptsize(82--104)} & 0.98 & 608\,{\scriptsize(608--608)} & 0.99 \\
 \bottomrule
 \end{tabular}
\end{table}

\paragraph{Within-family patterns differ by task.} Qwen3 endpoints shrink with model size on agreement ($1{,}054 \to 205 \to 190$ edges) and induction ($404 \to 228 \to 88$, at accuracies $0.94$, $0.98$, $0.98$), but not on IOI or docstring. Llama goes the other way on two tasks; agreement grows from 263 edges at 1B to 314 at 3B. The retained share falls from a median $7.5\%$ in GPT-2 to $0.37\%$ in Qwen3-4B, but the association across \NCells{} task--model cells is not significant (Spearman $-0.24$, $p=0.18$). Agreement accuracy also ranges from $0.71$ to $0.84$, so these sizes are only approximately performance-matched.

\paragraph{Why we report interacting partners per edge.} Most strong pair effects are sub-additive: $87\%$ satisfy $d_{ij}<d_i+d_j$. One edge can consume much of the available logit margin, leaving less effect for another to add. The fraction of pairs crossing a fixed $0.05$ interaction threshold also falls sharply as circuit size grows (Spearman $-0.89$). We report interacting partners per edge rather than compare those fractions across differently sized circuits.

\section{Scope: Ablation Convention and Where the Result Is Weakest}\label{app:ablation}

We use interchange ablation for every circuit: a closed edge reads the activation from the corrupted prompt. Mean and zero ablation can change accuracy substantially. Under mean ablation, docstring circuits on Gemma-3-1B at $K=64$ and Llama-3.2-3B at $K=512$ score $0.00$, while docstring/Llama-3.2-1B still scores $0.92$ at $K=512$. Matched-accuracy frozen circuits show similar sensitivity when they close a comparable fraction of edges. Frozen circuits that withstand other rules generally keep most of the graph. This pattern is consistent with sensitivity increasing as more edges are ablated; \citet{miller2024} also report sensitivity to the intervention convention. Our size comparisons use the same convention in every arm. The reported edge count does not assert that the circuit works when all other activity is set to zero.

\begin{table}[htbp]
\centering\small
\setlength{\tabcolsep}{6pt}
\caption{\textbf{Verification checks by task} (\S\ref{sec:e4}). Values are medians over eight models and three seeds. \emph{Accuracy after cutting} compares removal of all $C_1$ edges with an equal-size random removal. \emph{Load-bearing alone} is the share of edges that matter individually. \emph{No smaller subset} counts enumerable cells with no smaller faithful subset among all $2^k$ possibilities.}
\label{tab:pertask}
\begin{tabular}{@{}lccccc@{}}
\toprule
& \multicolumn{2}{c}{Accuracy after cutting} & \multicolumn{2}{c}{Load-bearing alone} & No smaller \\
\cmidrule(lr){2-3}\cmidrule(lr){4-5}
Task & $C_1$ & random edges & $C_1$ & frozen & subset \\
\midrule
IOI       & $0.52$ & $0.98$ & $15\%$ & $0\%$ & 2 of 7 \\
Agreement & $0.22$ & $0.62$ & $4\%$  & $0\%$ & none enumerable \\
Induction & $0.00$ & $0.96$ & $12\%$ & $0\%$ & 6 of 6 \\
Docstring & $0.00$ & $0.97$ & $3\%$  & $0\%$ & 3 of 6 \\
\bottomrule
\end{tabular}
\end{table}

\subsection{Where the Result Is Weakest}\label{app:weakest}

The condensed IOI circuit omits \RoleDropped{} of the \WangHeads{} heads mapped by \citet{wang2023}; it is sufficient for the task, but not a full reconstruction of their mechanism. Agreement accuracy ranges from $0.71$ to $0.84$, and docstring/GPT-2 reaches only $0.44$; we exclude that cell from the payoff analyses in Sections~\ref{sec:e4} and~\ref{sec:e5}. $C_1$ is also larger than frozen EAP-IG on docstring/Qwen3-0.6B and agreement/GPT-2. Several Gemma runs reach or approach the controller's edge floor, so we cannot tell how much further they might compress.

\section{Adapter Placement and Complexity}\label{app:lora}

Edge counts exclude the LoRA adapter, so Table~\ref{tab:lora} reports its size for one IOI endpoint per family. Every run uses rank $16$, $\alpha=32$, and no dropout. GPT-2 adapts fused attention and MLP projections; the other families adapt seven attention and MLP projection matrices per layer. The adapter spans the model but adds only $0.9$--$1.9\%$ as many parameters as the base model. Its updates are uneven: the top quarter of layers carries $38$--$61\%$ of the squared Frobenius norm of $BA$, and the largest individual updates occur in MLP projections in every family. These measurements locate weight changes; the ablation and faithfulness tests assess what the retained edges carry.

\begin{table}[htbp]
\centering
\small
\caption{\textbf{LoRA accounting for one representative endpoint per family} (IOI, seed 22). ``Top-quartile share'' is the fraction of total squared update norm carried by the top quarter of layers.}
\label{tab:lora}
\begin{tabular}{lrrrr}
\toprule
Model & Adapted matrices & LoRA params & \% of base & Top-quartile share \\
\midrule
GPT-2 small & 48 & 2.36M & 1.90\% & 37.6\% \\
Llama-3.2-1B & 112 & 11.27M & 0.91\% & 50.0\% \\
Gemma-3-1B & 182 & 13.05M & 1.30\% & 47.9\% \\
Qwen3-1.7B & 196 & 17.43M & 1.01\% & 61.3\% \\
\bottomrule
\end{tabular}
\end{table}

\section{Scope and Boundary Conditions}\label{app:limitations}

\paragraph{What circuit size measures.} $C_1$ describes the adapted model; its outputs closely track, but do not exactly match, the original (\S\ref{sec:e5}). Edge counts use interchange ablation throughout and exclude computation within retained heads and MLPs and the LoRA adapter. Both condensed and comparably sparse frozen circuits can fail under other ablation rules \citep{miller2024} (Appendices~\ref{app:ablation} and~\ref{app:lora}). Edge and node counts also differ: the per-task medians are \EdgesIOI{} edges over \NodesIOI{} nodes on IOI, \EdgesInd{} over \NodesInd{} on induction, \EdgesAgr{} over \NodesAgr{} on agreement, and \EdgesDoc{} over \NodesDoc{} on docstring. Even the 3-edge Gemma-3-1B endpoints span \NodesSmallest{} nodes. On IOI, the circuit is sufficient for the behavior without recovering every part of the published mechanism (Appendix~\ref{app:weakest}).

\paragraph{How endpoints are selected.} The tolerance, cut schedule, 500-step healing ceiling, and LoRA rank were fixed before the grid, so each returned circuit meets the same validation criteria. These settings do not guarantee the smallest faithful circuit: one 12-edge endpoint contains a faithful 3-edge subset, and we did not sweep the initial cut fraction or its decay. The ordering persists under a $90\%$-accuracy selection rule (Appendix~\ref{app:budget}); shorter pruning routes fail at endpoints of 16 edges or fewer despite four times one round's healing budget (Table~\ref{tab:path}). Joint gate-and-adapter training returns admissible circuits in 12 of \WThreeCells{} cells, with \WThreeSmaller{} smaller than $C_1$; the controller returns a valid circuit across the grid without per-cell sparsity tuning (Appendix~\ref{app:joint}).

\paragraph{When the capability constraint binds.} On docstring, accuracy stays at $1.000$ until the capability gate stops pruning; on agreement, it can improve as the circuit shrinks. Those endpoints are valid circuits, not measured global minima. The gate rarely rejects Gemma cuts because healing improves probe perplexity, compared with $15$--$20\%$ of proposals in the other families (Appendix~\ref{app:capability}). Excluding Gemma still leaves a \ShrinkNoGemma{} average reduction, with $C_1$ smaller in \ShrinkNoGemmaCells{} of 24 cells. We did not calibrate the gate separately by model, and the generic-text probe misses some nearby task damage: harder IOI prompts degrade in three of four families.

\paragraph{Cost and transfer.} Condensation takes more computation than frozen discovery; the cost is most useful when the circuit is reused or enables an otherwise infeasible analysis (Appendix~\ref{app:cost}).

Three seeds measure optimization stability: endpoints have median containment \SeedContainment{} per task and edge overlap \SeedNullMin{}--\SeedNullMax{} above a matched-size null (Appendix~\ref{app:seedstab}). We have not tested whether the same circuit appears under other checkpoints, input distributions, graph granularities, or pruning procedures; prior work reports sensitivity to these choices \citep{bali2026headstability,rai2026,parekh2026circus}.

\section{Containment in the Frozen Ranking}\label{app:containment}

We rank the original model's edges with EAP-IG on training data, then select the smallest frozen circuit within $0.05$ of its full-graph validation accuracy. We match edges by name across this circuit and $C_1$ and count how many $C_1$ edges it contains. For a random set of $C_1$'s size, the expected overlap is $k_{\text{sel}}/|E|$: a frozen circuit holding one third of the graph would contain one third of a random edge set on average.

Median containment across \NCells{} cells is \ContainMed{}, versus \ContainNull{} expected by chance; every cell exceeds its own chance level. In two cells, $C_1$ is larger than the frozen circuit, so full containment is impossible. At seed 11, agreement/GPT-2 has 595 versus 343 edges, and docstring/Qwen3-0.6B has 1{,}976 versus 798. Their containment scores are $0.57$ and $0.39$, still three to seven times chance. Even to include every $C_1$ edge, the frozen ranking extends no further than its top \ContainTopFrac{}.

The overlap shows that condensation uses edges ranked highly in the original model. It cannot establish that they perform the same computation after their weights change; the behavioral comparisons in \S\ref{sec:e5} address that separate question.

\section{Post-Training Arms}\label{app:posttrain}

Section~\ref{sec:e2} uses supervised fine-tuning as the post-training comparison. We also tried DPO, GRPO, and REINFORCE, but the short-answer tasks give the reinforcement arms little usable reward variation.

Each arm trains a LoRA adapter on the task using the same data splits and head-level graph as the other arms. Table~\ref{tab:posttrain} reports the EAP-IG participation ratio after training divided by its value before, so values below $1$ indicate concentrated attribution. It also reports task accuracy and the general-probe perplexity ratio used by the controller's capability gate.

\begin{table}[H]
\centering\small
\caption{\textbf{Post-training arms.} Ratios are after/before, with values below $1$ indicating concentration. \emph{Probe} is general-corpus perplexity relative to the original model, and the controller's limit is $1.05$. DPO covers fewer cells because induction and docstring do not provide its required contrast answer.}
\label{tab:posttrain}
\begin{tabular}{@{}lccccc@{}}
\toprule
Arm & Cells & PR ratio & Concentrates & Task acc. & Probe \\
\midrule
SFT        & $32$ & $0.76$ & $23/32$ & $0.99 \to 1.00$ & $1.37\times$ \\
DPO \citep{rafailov2023dpo} & $16$ & $0.51$ & $15/16$ & $0.89 \to 1.00$ & $12.9\times$ \\
GRPO \citep{shao2024} & $10$ & $0.72$ & $10/10$ & $0.76 \to 0.82$ & $1.0\times$ \\
REINFORCE \citep{williams1992reinforce} & $10$ & $1.88$ & $3/10$  & $0.76 \to 0.49$ & $6{,}336\times$ \\
\bottomrule
\end{tabular}
\end{table}

In 16 evaluated cells, the mean reward is \RlRewardLow{} (at most $0.022$): sampled answers rarely match the target. In the other four it averages \RlRewardHigh{}, so almost every sample matches. Both regimes give little reward variation. Moreover, these rewards use exact match on generated text, while the paper's accuracy metric uses restricted choice. REINFORCE falls from $0.76$ to $0.49$ task accuracy and raises probe perplexity by $6{,}336\times$. GRPO stays inside the capability budget but improves accuracy only from $0.76$ to $0.82$ (Table~\ref{tab:posttrain}).

These runs do not provide a matched reinforcement comparison. A different reward and evaluation objective would be needed before comparing RL post-training with condensation on these tasks.

\end{document}